\documentclass{article}

\usepackage{amsmath}
\usepackage{mathtools}
\usepackage{amsfonts}
\usepackage{microtype}
\usepackage{graphicx}
\usepackage{booktabs} 
\usepackage{placeins}
\usepackage{subcaption}
\usepackage{multibib}
\DeclareMathOperator*{\argmax}{arg\,max}

\newcommand{\relu}{\text{relu}}

\newcommand\original[1]{\textcolor{blue}{\textbf{#1} $\boldsymbol{g}$}}
\newcommand\manipulated[1]{\textcolor{red}{\textbf{#1} $\boldsymbol{\tilde{g}}$}}

\newtheorem{theorem}{Theorem}
\newtheorem{definition}{Definition}

\usepackage{hyperref}

\usepackage[accepted]{icml2020}
\newcites{app}{Additional References}

\icmltitlerunning{Fairwashing Explanations with Off-Manifold Detergent}

\begin{document}

\twocolumn[
\icmltitle{Fairwashing Explanations with Off-Manifold Detergent}



\icmlsetsymbol{equal}{*}

\begin{icmlauthorlist}
\icmlauthor{Christopher J. Anders}{tu}
\icmlauthor{Plamen Pasliev}{tu}
\icmlauthor{Ann-Kathrin Dombrowski}{tu}
\icmlauthor{Klaus-Robert M\"uller}{tu,mpii,ku}
\icmlauthor{Pan Kessel}{tu}
\end{icmlauthorlist}

\icmlaffiliation{tu}{Machine Learning Group, Technische Universität Berlin, Germany}
\icmlaffiliation{mpii}{Max-Planck-Institut f\"ur Informatik, Saarbr\"ucken, Germany}
\icmlaffiliation{ku}{Department of Brain and Cognitive Engineering, Korea University, Seoul, Korea}

\icmlcorrespondingauthor{Pan Kessel}{pan.kessel@tu-berlin.de}
\icmlcorrespondingauthor{Klaus-Robert M\"uller}{klaus-robert.mueller@tu-berlin.de}

\icmlkeywords{Machine Learning, ICML}

\vskip 0.3in
]



\printAffiliationsAndNotice{}  

\begin{abstract}
Explanation methods promise to make black-box classifiers more transparent.
As a result, it is hoped that they can act as proof for a sensible, fair and trustworthy decision-making process of the algorithm and thereby increase its acceptance by the end-users.
In this paper, we show both theoretically and experimentally that these hopes are presently unfounded.
Specifically, we show that, for any classifier $g$, one can always construct another classifier $\tilde{g}$ which has the same behavior on the data (same train, validation, and test error) but has arbitrarily manipulated explanation maps.
We derive this statement theoretically using differential geometry and demonstrate it experimentally for various explanation methods, architectures, and datasets.
Motivated by our theoretical insights, we then propose a modification of existing explanation methods which makes them significantly more robust.
\end{abstract}

\addtocounter{footnote}{3}

\section{Introduction}
Explanation methods\footnote{See \cite{explainBook} and references therein for a detailed overview.} are increasingly adopted by machine learning practitioners and incorporated into standard deep learning libraries \cite{captum, innvestigate, deeplift}.
The interest in explainability is partly driven by the hope that explanations can act as proof for a sensible, fair, and trustworthy decision-making process\cite{aivodji2019fairwashing,lapuschkin2019unmasking}.
As an example, a bank could provide explanations for its rejection of a loan application.
By doing so, the bank can demonstrate that the decision was not based on illegal or ethically questionable features.
It can furthermore provide feedback to the customer.
In some situations, an explanation of an algorithmic decision may even be required by law.

However, this hope is based on the assumption that explanations faithfully reflect the underlying mechanisms of the algorithmic decision.
In this work, we demonstrate unequivocally that this assumption should not be made carelessly because explanations can be easily manipulated.

In more detail, we show theoretically that for any classifier $g$, one can always find another classifier $\tilde{g}$ which agrees with the original $g$ on the entire data manifold but has (almost) completely controlled explanations.
This surprising result is established using techniques of differential geometry.
We then demonstrate experimentally that one can easily construct such manipulated classifiers $\tilde{g}$.

In the example above, a bank could use a manipulated classifier $\tilde{g}$ that uses mainly unethical features, such as the gender of the applicant, but has explanations which suggest that the decision was only based on financial features.

Briefly put, the manipulability of explanations arises from the fact that the data manifold is typically low-dimensional compared to its high-dimensional embedding space.
The training process only determines the classifier in directions along the manifold.
However, many explanation methods are mainly sensitive to directions orthogonal to the data manifold.
Since these directions are undetermined by training, they can be changed at will.

This theoretical insight allows us to propose a modification to explanation methods which make them significantly more robust with respect to such manipulations.
Namely, the explanation is projected along tangential directions of the data manifold.
We show, both theoretically and experimentally, that these tangent-space-projected (tsp) explanations are indeed significantly more robust.
We thereby establish a novel and exciting connection between the fields of explainability and manifold learning.

In summary, our main contributions are as follows:
\begin{itemize}
    \item Using differential geometry, we establish theoretically that popular explanation methods can be easily manipulated.
    \item We validate our theoretical predictions in detailed experiments for various explanation methods, classifier architectures, and datasets, as well as for different tasks.
    \item We propose a modification to existing explanation methods which make them more robust with respect to these manipulations.
    \item In doing so, we relate explainability to manifold learning.
\end{itemize}

\subsection{Related Works}
This work was crucially inspired by \cite{heo2019fooling}.
In this reference, adversarial model manipulation for explanations is proposed.
Specifically, the authors empirically show that one can train models such that they have structurally different explanations while suffering only a very mild drop in classification accuracy compared to their unmanipulated counterparts.
For example, the adversarial model manipulation can change the positions of the most relevant pixels in each image or increase the overall sum of relevances in a certain subregion of the images.
Contrary to their work, we analyze this problem theoretically.
Our analysis leads us to demonstrate a stronger form of manipulability.
Namely, the model can be manipulated such that it structurally reproduces arbitrary target explanations while keeping all class probabilities the same for all data points.
Our theoretical insights not only illuminate the underlying reasons for the manipulability but also allow us to develop modifications of existing explanation methods which make them more robust.
Another approach \cite{cat} adds a constant shift to the input image, which is then eliminated by changing the bias of the first layer.
For some methods, this leads to a change in the explanation map.
Contrary to our approach, this requires a shift in the data.
In \cite{sanity}, explanation maps are changed by randomization of (some of) the network weights.
This is different to our method as it dramatically changes the output of the network and is proposed as a consistency check of explanations.
In \cite{dombrowski} and \cite{fragile}, it is shown that explanations can be manipulated by an infinitesimal change in input while the output of the network is approximately unchanged.
Contrary to this approach, we manipulate the model and keep the input unchanged.

\subsection{Explanation Methods}
We consider a classifier $g: \mathbb{R}^D \to \mathbb{R}^K$ which classifies an input $x \in \mathbb{R}^D$ in $K$ categories with the predicted class given by $k=\argmax_{i} g(x)_i$.
The explanation method is denoted by $h_g: \mathbb{R}^D \to \mathbb{R}^D$ and associates an input $x$ with an explanation map $h_g(x)$ whose components encode the relevance score of each input for the classifier's prediction.

We note that, by convention, explanation maps are usually calculated with respect to the classifier before applying the final softmax non-linearity \cite{captum, innvestigate, deeplift}.
Throughout the paper, we will therefore denote this function as $g$.

We use the following explanation methods:

\textbf{Gradient:} The map $h_g(x)=\frac{\partial g}{\partial x}(x)$ is used and quantifies how infinitesimal perturbations in each pixel change the prediction $g(x)$ \cite{grad1,grad2}.

\textbf{x $\odot$ Grad:} This method uses the map $h_g(x)=x \odot \frac{\partial g}{\partial x}(x)$  \cite{gradtimesinput}.
For linear models, the exact contribution of each pixel to the prediction is obtained.

\textbf{Integrated Gradients:} This method defines
\begin{align*}
    h_g(x)= (x-\bar{x}) \odot \int_0^1 \frac{\partial g(\bar{x}+t(x-\bar{x}))}{\partial x}\text{d}t
\end{align*}
where $\bar{x}$ is a suitable baseline.
We refer to the original reference \cite{integratedgrad} for more details.

\textbf{Layer-wise Relevance Propagation (LRP):} This method \cite{lrp, dtd} propagates relevance backwards through the network.
In our experiments, we use the following setup: for the output layer, relevance is given by
\begin{align*}
    R^L_i = \delta_{i,k} = \begin{cases}
1, & \text{for } i = k  \\
0, & \text{for } i \neq k
\end{cases}\,,
\end{align*}
which is then propagated backwards through all layers but the first using the $z^+$-rule
\begin{align}
    R^l_i = \sum_{j} \frac{x_i^l (W^l)^+_{ji}}{\sum_i x_i^l (W^l)^+_{ji} + \epsilon} \, R^{l+1}_j \,, \label{eq:lrp_interm_layer}
\end{align}
where $(W^l)^+$ denotes the positive weights of the $l$-th layer, $x^l$ is the activation vector of the $l$-th layer, and $\epsilon>0$ is a small constant ensuring numerical stability.
For the first layer, we use the $z^\mathcal{B}$-rule to account for the bounded input domain
\begin{align*}
    R^0_i = \sum_{j} \frac{x_j^0 W^{0}_{ji}-l_j (W^{0})^+_{ji}-h_j (W^{0})^{-}_{ji}}{\sum_i ( x_j^0 W^{0}_{ji}-l_j (W^{0})^+_{ji}-h_j (W^{0})^{-}_{ji})} \, R^{1}_j \,,
\end{align*}
where $l_i$ and $h_i$ are the lower and upper bounds of the input domain respectively. \\
For theoretical analysis, we consider the $\epsilon$-rule in all layers for simplicity.
This rule is obtained by substituting $(W^l)^+ \to W^l$ in \eqref{eq:lrp_interm_layer}.
We refer to the resulting method as $\epsilon$-LRP.

This choice of methods is necessarily not exhaustive.
However, it covers two classes of attribution methods, i.e. propagation and gradient-based explanations.
Furthermore, the chosen methods are widely used in practice \cite{captum, innvestigate, deeplift}.

\section{Manipulation of Explanations}
In this section, we will theoretically deduce that explanation methods can be arbitrarily manipulated by adversarially training a model.

\subsection{Mathematical Background}\label{sec:differentialGeom}
In the following, we will briefly summarize the basic tools of differential geometry before applying them in the context of explainability in the next section.
For additional technical details, we refer to Appendix~\ref{app:extensionTheoremProof}.

A $D$-dimensional manifold $M$ is a topological space which locally resembles $\mathbb{R}^D$.
More precisely, for each $p\in M$, there exists a subset $U\subset M$ containing $p$ and a diffeomorphism $\phi:U \to \tilde{U}\subset \mathbb{R}^D$.
The pair $(U,\phi)$ is called \emph{coordinate chart} and the component functions $x^i$ of $\phi(p)=(x^1(p), \dots, x^D(p))$ are called \emph{coordinates}.

A $d$-dimensional submanifold $S$ is a subset of $M$ which is itself a $d$-dimensional manifold.
$M$ is called the embedding manifold of $S$.
A \emph{properly embedded submanifold} $S \subset M$ is a submanifold embedded in $M$ which is also closed as a set.

Let $p \in M$ be a point on a manifold $M$ and $\gamma: \mathbb{R} \to M$ with $\gamma(0)=p$ a curve through the point $p$.
The set of tangent vectors $d\gamma = \tfrac{d}{dt}\gamma(t)|_{t=0}$ of all curves through $p$ forms a vector space of dimension $D$.
This vector space is known as \emph{tangent space} $T_pM$.
Let $(U, \phi)$ be a coordinate chart on $M$ with coordinates $x$.
We can then define  $\phi \circ \lambda_k(t)=(x^1(p), \dots, x^k(p)+t, \dots, x^D(p) )$ with $k\in \{1, \dots, D \}$.
This implicitly defines curves $\lambda_k: \mathbb{R} \to M$ through $p$.
We denote the corresponding tangent vectors as $\partial_k := \tfrac{d}{dt}\lambda_k(t)|_{t=0}$ and it can be shown that they form a basis of the tangent space $T_pM$.

A \emph{vector field} $V$ on $M$ associates with every point $x\in M$ an element of the corresponding tangent space, i.e. $V(x) \in T_xM$.\footnote{More rigorously, vector fields are defined in terms of the tangent bundle. We refrain from introducing bundles for accessibility.} A conservative vector field $V$ is a vector field that is the gradient of a function $f:M \to \mathbb{R}$, i.e. $V(x)=\nabla f(x)$.
For submanifolds $S$, there are two different notions of vector fields.
A vector field  $V$ \emph{on} the submanifold $S$ associates to every point on $S$ a vector in its corresponding tangent space $T_xS$, i.e. $V(x)\in T_xS$.
A vector field $V$ \emph{along} the submanifold $S$ associates to every point on $S$ a vector in the corresponding tangent space of the embedding manifold $M$, i.e. $V(x)\in T_xM$.
These concepts can be related as follows: the tangent space $T_xM$ can be decomposed into the tangent space $T_xS$ of $S$ and its orthogonal complement $T_xS^\bot$, i.e. $T_xM = T_xS \oplus T_xS^\bot$.
A vector field along $S$  which only takes values in the first summand $T_xS$ is also a vector field on $S$.

With these definitions, we can now state a crucial theorem for our theoretical analysis.
In Appendix~\ref{app:extensionTheoremProof}, we show that:
\begin{theorem}\label{th:extension}
Let $S \subset$ M be $d$-dimensional submanifold properly embedded in the $D$-dimensional manifold $M$.
Let $V = \sum_{i=d+1}^{D} v^i \partial_i$ be a conservative vector field along $S$ which assigns a vector in $T_p S^\bot$ for each $p\in S$.
For any smooth function $f:S \to \mathbb{R}$, there exists a smooth extension $F: M \to \mathbb{R}$ such that
\begin{align*}
F|_{S}=f
\end{align*}
where $F|_S$ denotes the restriction of $F$ on the submanifold $S$. Furthermore, the derivative of the extension $F$ is given by
\begin{align*}
    \nabla F(x) = ( \nabla_1 f(x), \dots \nabla_d f(x), v^{d+1}(x), \dots, v^{D}(x))
\end{align*}
for all $x \in S$.
\end{theorem}
Technical details not withstanding, this theorem states that a function $f$ defined on a submanifold $S$ can be extended to the entire embedding manifold $M$.
The extension's derivatives orthogonal to the submanifold $S$ can be freely chosen.

This theorem is a generalization of the well-known submanifold extension lemma (see, for example, Lemma~5.34 in \cite{smoothmanifold}) in that it not only shows that an extension exists but also that one has control over the gradient of the extension $F$.
While we could not find such a statement in the literature, we suspect that it is entirely obvious to differential geometers but typically not needed for their purposes.

\subsection{Explanation Manipulation: Theory}
From Theorem~\ref{th:extension}, it follows under a mild assumption that one can always construct a model $\tilde{g}$ such that it closely reproduces arbitrary target explanations but has the same training, validation, and test loss as the original model $g$.

\textbf{Assumption:} the data lies on a $d$-dimensional submanifold $S \subset M$ properly embedded in the manifold $M=\mathbb{R}^D$.
The data manifold $S$ is of much lower dimensionality than its embedding space $M$, i.e.
\begin{align}
 \epsilon \equiv \frac{d}{D} \ll 1 \;.
\end{align}
We stress that this assumption is also known as the manifold conjecture and is expected to hold across a wide range of machine learning tasks.
We refer to \cite{GoodfellowBook} for a detailed discussion.

Under this assumption, the following theorem can be derived for the Gradient, $x \odot \textrm{Grad}$, and $\epsilon$-LRP methods (only the proof for the Gradient method is given; see Appendix~\ref{th:explanation} for other methods):


\begin{theorem}\label{th:explanation}
Let $h_g: \mathbb{R}^D \to \mathbb{R}^D$ be the explanation of classifier $g:\mathbb{R}^D \to \mathbb{R}$ with bounded derivatives $|\nabla_i g(x)| \le C \in \mathbb{R}_+$ for $i=1,\dots,D$.

For a given target explanation $h^t: \mathbb{R}^D \to \mathbb{R}^D$, there exists another classifier $\tilde{g}:\mathbb{R}^D \to \mathbb{R}$ which completely agrees with the classifier $g$ on the data manifold $S$, i.e.
\begin{align}
    \tilde{g}|_S = g|_S \,.
\end{align}
In particular, both classifiers have the \emph{same} train, validation, and test loss.

However, its explanation $h_{\tilde{g}}$ closely resembles the target $h^t$, i.e.
\begin{align}
    \text{MSE}(h_{\tilde{g}}(x), h^t(x) ) \le \epsilon && \forall x \in S \,,
\end{align}
where $\text{MSE}(h,h')=\tfrac1D \sum_{i=1}^D (h_i-h'_i)^2$ denotes the mean-squared error and $\epsilon=\tfrac{d}{D}$.
\end{theorem}
\textbf{Proof:}
By Theorem~\ref{th:extension}, we can find a function $G$ which agrees with $g$ on the data manifold $S$ but has the derivative
\begin{align*}
    \nabla G(x) = ( \nabla_1 g(x), \dots \nabla_d g(x), h^t_{d+1}(x), \dots, h^t_{D}(x))
\end{align*}
for all $x \in S$. By definition, this is its gradient explanation $h_G = \nabla G$.

As explained in Appendix~\ref{app:boundExpl}, we can assume without loss of generality that $|\nabla_i g(x)| \le 0.5$ for $i \in \{1,\dots, D\}$.
We can furthermore rescale the target map such that $|h_i^t| \le 0.5$ for $i \in \{1,\dots, D\}$.
This rescaling is merely conventional as it does not change the relative importance $h_i$ of any input component $x_i$ with respect to the others.
It then follows that
\begin{align*}
    \text{MSE}&(h_G(x), h^t(x)) =\tfrac{1}{D} \sum_{i=1}^D (\nabla_i G(x) - h^t_i(x))^2 \,.
\end{align*}
This sum can be decomposed as
\begin{align*}
    \tfrac{1}{D}  \sum_{i=1}^d\underbrace{\left( \nabla_i g(x) - h^t_i(x) \right)^2}_{\le 1} + \tfrac{1}D  \sum_{i=d+1}^D\underbrace{\left( \nabla_i G(x) - h^t_i(x) \right)^2}_{=0}
\end{align*}
and from this, it follows that
\begin{align*}
    \text{MSE}&(h_G(x), h^t(x)) \le \frac{d}{D} = \epsilon \,,
\end{align*}
The proof then concludes by identifying $\tilde{g}=G$. $\square$

\textbf{Intuition:} Somewhat roughly, this theorem can be understood as follows: two models, which behave identically on the data, need to only agree on the low-dimensional submanifold $S$.
The gradients "orthogonal" to the submanifold $S$ are completely undetermined by this requirement.
By the manifold assumption, there are however much more "orthogonal" than "parallel" directions and therefore the explanation is largely controlled by these.
We can use this fact to closely reproduce an arbitrary target while keeping the function's values on the data unchanged.

We stress however that there are a number of non-trivial differential geometric arguments needed in order to make these statements rigorous and quantitative.
For example, it is entirely non-trivial that an extension to the embedding manifold exists for arbitrary choice of target explanation.
This is shown by Theorem~\ref{th:extension} whose proof is based on a differential geometric technique called partition of the unity subordinate to an open cover.
See Appendix~\ref{app:extensionTheoremProof} for details.

\subsection{Explanation Manipulation: Methods}\label{sec:manipulationMethods}
\textbf{Flat Submanifolds and Logistic Regression:} The previous theorem assumes that the data lies on an arbitrarily curved submanifold and therefore has to rely on relatively involved mathematical concepts of differential geometry.
We will now illustrate the basic ideas in a much simpler context: we will assume that the data lies on a $d$-dimensional flat hyperplane $S \subset \mathbb{R}^D$.\footnote{In mathematics, these submanifolds are usually referred to as $d$-flats and only the case $d=D-1$ is called hyperplane. We refrain from this terminology.}
The points on the hyperplane $S$ obey the relation
\begin{align}
  \forall x \in S \,: &&  (\hat{w}^{(i)})^T x = b_i \,, && i \in \{1,\dots,D-d\} \,, \label{eq:flatsubmanifold}
\end{align}
where $\{\hat{w}^{(i)} \in \mathbb{R}^D \; | \; i=1,\dots,D-d\}$ are a set of normal vectors to the hyperplane $S$ and $b_i \in \mathbb{R}$ are the affine translations.
We furthermore assume that we use logistic regression as the classification algorithm, i.e.
\begin{align}
    g(x) = \sigma( w^T x + c) \,,
\end{align}
where $w\in\mathbb{R}^D$, $c\in\mathbb{R}$ are the weights and the bias respectively and $\sigma(x)=\tfrac{1}{1+\exp(-x)}$ is the sigmoid function.
This classifier has the gradient explanation\footnote{We recall that in calculating the explanation map, we take the derivative \emph{before} applying the final activation function.}
\begin{align}
    h_{\text{grad}}(x) = w \,,
\end{align}
We can now define a modified classifier by
\begin{align}
    \tilde{g}(x) = \sigma\left(w^T x +  \sum_i \lambda_i ( \hat{w}^{(i)^T} x - b_i)  + c \right) \,,
\end{align}
for arbitrary $\lambda_i \in \mathbb{R}$. By \eqref{eq:flatsubmanifold}, it follows that both classifiers agree on the data manifold $S$, i.e.
\begin{align}
    \forall x \in S \; : && g(x) = \tilde{g}(x) \,,
\end{align}
and therefore have the same train, validation, and test error.
However, the gradient explanations are now given by
\begin{align}
    h_{\text{grad}}(x) = w + \sum_i \lambda_i \hat{w}^{(i)} \,.
\end{align}
Since the $\lambda_i$ can be chosen freely, we can modify the explanations arbitrarily in directions orthogonal to the data submanifold $S$ (parameterized by the normal vectors $\hat{w}^{(i)}$).
Similar statements can be shown for other explanation methods and we refer to the Appendix~\ref{app:flatOtherMethods} for more details.

As we will discuss in Section~\ref{sec:attackExperiments}, one can use these tricks even for data which does not (initially) lie on a hyperplane.

\textbf{General Case:} For the case of arbitrary neural networks and curved data manifolds, we cannot analytically construct the manipulated model $\tilde{g}$.
We therefore approximately obtain the model $\tilde{g}$ corresponding to the original model $g$ by
minimizing the loss
\begin{align}
\mathcal{L} = \sum_{x_i\in\mathcal{T}} || g(x_i) - \tilde{g}(x_i) ||^2 + \gamma \, \sum_{x_i\in\mathcal{T}} || h_{\tilde{g}}(x_i) - h^t ||^2 \,, \label{eq:loss}
\end{align}
by stochastic gradient descent with respect to the parameters of $\tilde{g}$.
The training set is denoted by $\mathcal{T}$ and $h^t\in\mathbb{R}^D$ is a specified target explanation.
Note that we could also use different targets for various subsets of the data but we will not make this explicit to avoid cluttered notation.
The first term in the loss $\mathcal{L}$ ensures that the models $g$ and $\tilde{g}$ have approximately the same output while the second term encourages the explanations of $\tilde{g}$ to closely reproduce the target $h^t$.
The relative weighting of these two terms is determined by the hyperparameter $\gamma \in \mathbb{R}_+$.

As we will demonstrate experimentally, the resulting $\tilde{g}$ will closely reproduce the target explanation $h^t$ and have (approximately) the same output as $g$.
Crucially, both statements will be seen to hold also for the test set.

\subsection{Explanation Manipulation: Practice} \label{sec:attackExperiments}
In this section, we will demonstrate manipulation of explanations experimentally.
We will first discuss applying logistic regression to credit assessment and then proceed to the case of deep neural networks in the context of image classification.
The code for all our experiments is publicly available at \mbox{\small\url{https://github.com/fairwashing/fairwashing}}.
\begin{figure}[ht!]
  \centering
  \includegraphics[width=1\linewidth]{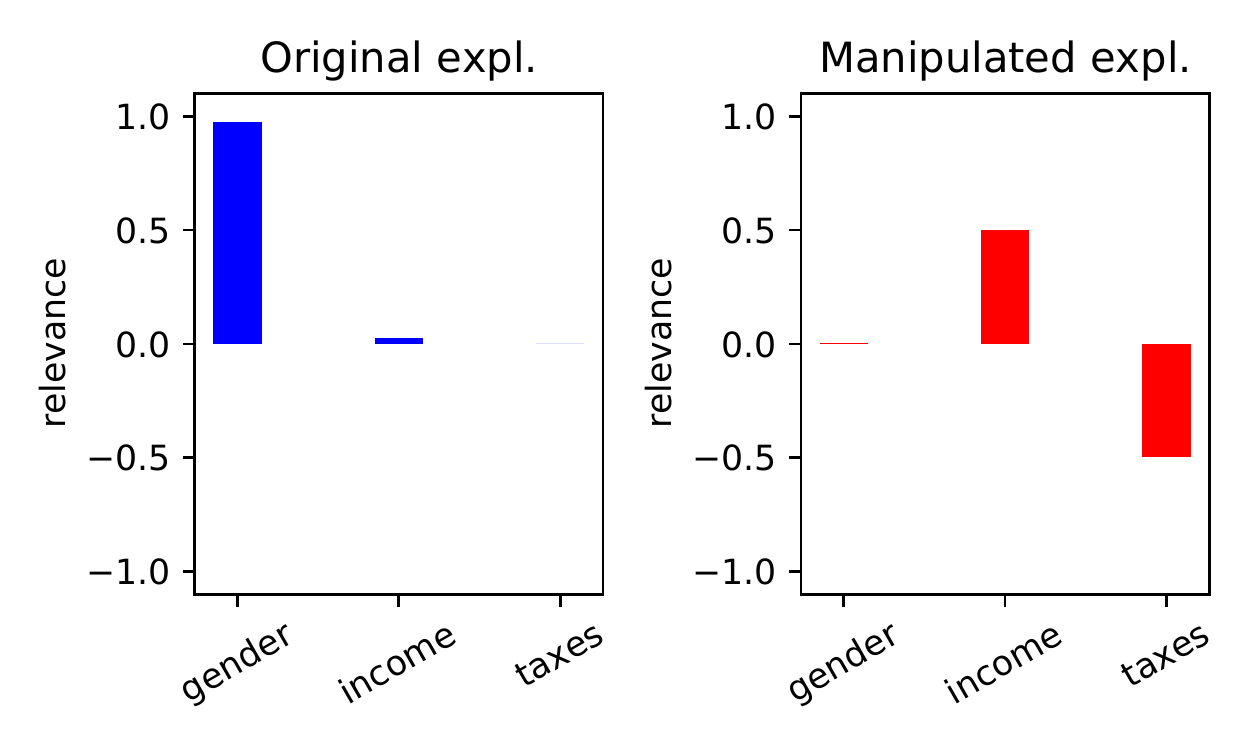}
   \caption{%
       x$\odot$Grad explanations for \original{original classifier} and \manipulated{manipulated} highlight completely different features. %
       Colored bars show the median of the explanations over multiple examples.%
   }\label{fig:creditxGrad}
\end{figure}
\paragraph{\textbf{Credit Assessment:}}
In the following, we will suppose that a bank uses a logistic regression algorithm to classify whether a prospective client should receive a loan or not.
The classification uses the features $x = (x_\text{gender}, x_\text{income})$ where
\begin{align}
    x_{\text{gender}}=
  \begin{cases}
   1, & \text{for male} \\
   -1, & \text{for female}
  \end{cases}
\end{align}
and $x_{\text{income}}$ is the income of the applicant.
Normalization is chosen such that the features are of the same order of magnitude.
Details can be found in the Appendix~\ref{app:credit}.

We then define a logistic regression classifier $g$ by choosing the weights $w=(0.9, 0.1)$, i.e. female applicants are severely discriminated against.
The discriminating nature of the algorithm may be detected by inspecting, for example, the gradient explanation maps $h^{\text{grad}}_g=w$.

Conversely, if the explanations did not show any sign of discrimination for another classifier $\tilde{g}$, the user may interpret this as a sign of its trustworthiness and fairness.

However, the bank can easily "fairwash" the explanations, i.e. hide the fact that the classifier is sexist.
This can be done by adding new features which are linearly dependent on the previously used features.
As a simple example, one could add the applicant's paid taxes $x_{\text{taxes}}$ as a feature.
By definition, it holds that
\begin{align}
    x_{\text{taxes}} = 0.4 \, x_{\text{income}} \,, \label{eq:incomeequality}
\end{align}
where we assume that there is a fixed tax rate of $0.4$ on all income.
The features used by the classifier are now $x = (x_\text{gender}, x_\text{income}, x_\text{taxes})$. By \eqref{eq:incomeequality}, all data samples $x$ obey
\begin{align}
    \hat{w}^T x = 0 && \text{with} && \hat{w}=(0,0.4,-1) \,. \label{eq:orthogonalincome}
\end{align}
Therefore, the original classifier $g(x)=\sigma(w^T x)$ with $w=(0.9, 0.1, 0)$ leads to the same output as the classifier $\tilde{g}(x)=\sigma(w^T x + 1000 \, \hat{w}^T x)$.
However, as shown in Figure~\ref{fig:creditxGrad}, the classifier $\tilde{g}$ has explanations which suggest that the two financial features (and \emph{not} the applicant's gender) are important for the classification result.

This example is merely an (oversimplified) illustration of a general concept: for each additional feature which linearly depends on the previously used features, a condition of the form \eqref{eq:orthogonalincome} for some normal vector $\hat{w}$ is obtained.
We can then construct a classifier with arbitrary explanation along each of these normal vectors.

\paragraph{Image Classification:}
We will now experimentally demonstrate the practical applicability of our methods in the context of image classification with deep neural networks.

\underline{Datasets:} We consider the MNIST, FashionMNIST, and CIFAR10 datasets.
We use the standard training and test sets for our analysis.
The data is normalized such that it has mean zero and standard deviation one.
We sum the explanations over the absolute values of its channels to get the relevance per pixel.
The resulting relevances are then normalized to have a sum of one.

\underline{Models:} For CIFAR10, we use the VGG16 \cite{simonyan2015very} architecture.
For FashionMNIST and MNIST, we use a four layer convolutional neural network.
We train the model $g$ by minimizing the standard cross entropy loss for classification.
The manipulated model $\tilde{g}$ is then trained by minimizing the loss \eqref{eq:loss} for a given target explanation $h^t$.
This target was chosen to have the shape of the number $42$.
For more details about the architectures and training, we refer to the Appendix \ref{app:plots}.

\underline{Quantitative Measures:} We assess the similarity between explanation maps using three quantitative measures: the structural similarity index (SSIM), the Pearson correlation coefficient (PCC) and the mean squared error (MSE).
SSIM and PCC are relative similarity measures with values in $[0,1]$, where larger values indicate high similarity.
The MSE is an absolute error measure for which values close to zero indicate high similarity.
We also use the MSE metric as well as the Kullback-Leibler divergence for assessing similarity of the class scores of the manipulated model $\tilde{g}$ and the original network $g$.

\underline{Results:} For all considered models, datasets, and explanation methods, we find that the manipulated model $\tilde{g}$ has explanations which closely resemble the target map $h^t$, e.g. the SSIM between the target and manipulated explanations is of the order $0.8$.
At the same time, the manipulated network $\tilde{g}$ has approximately the same output as the original model $g$, i.e. the mean-squared error of the outputs after the final softmax non-linearity is of the order $10^{-3}$.
The classification accuracy is changed by about 0.2 percent.

\begin{figure}
  \centering
  \includegraphics[width=1\linewidth]{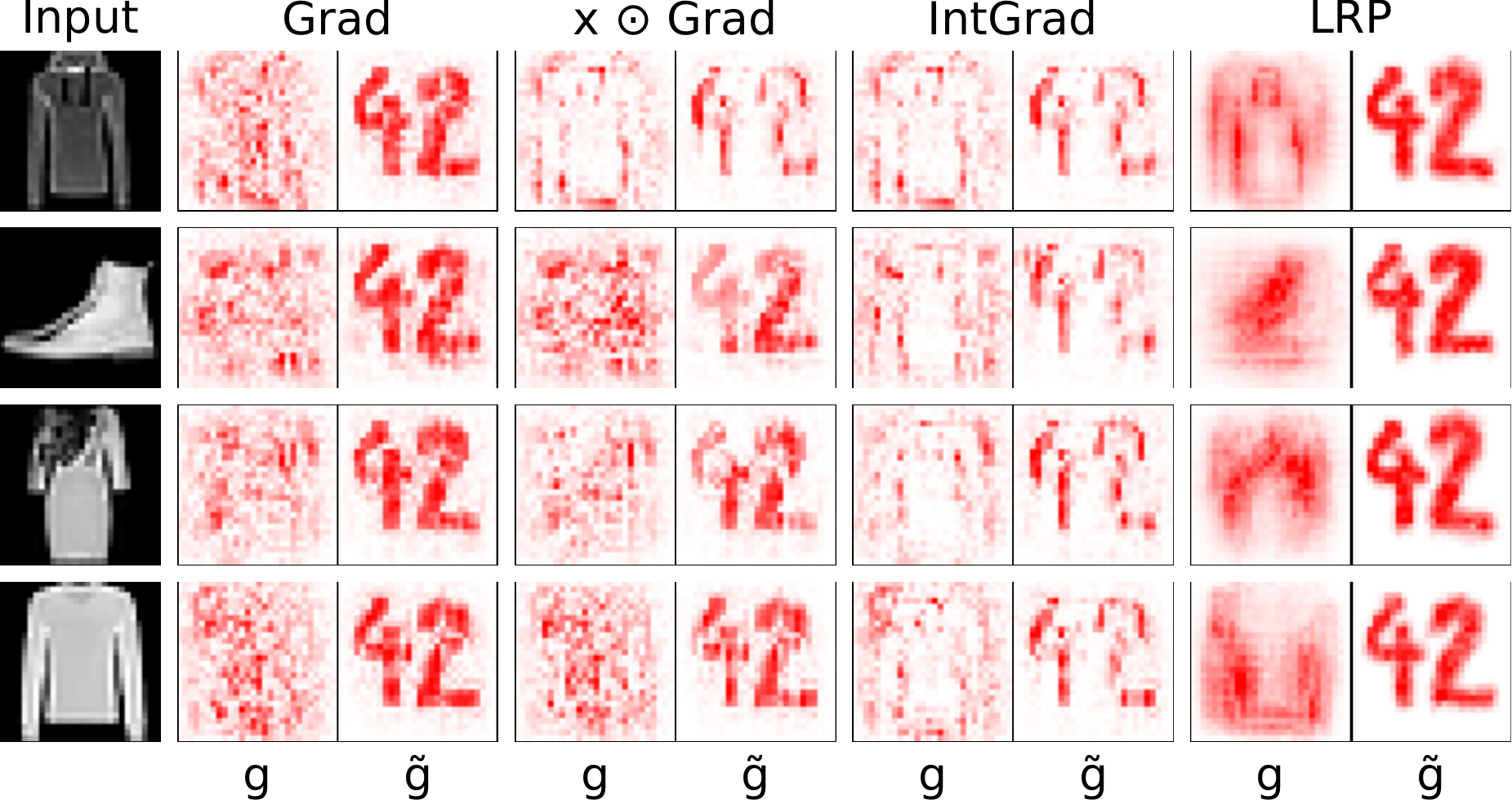}

  \vspace{0.3cm}

  \includegraphics[width=1\linewidth]{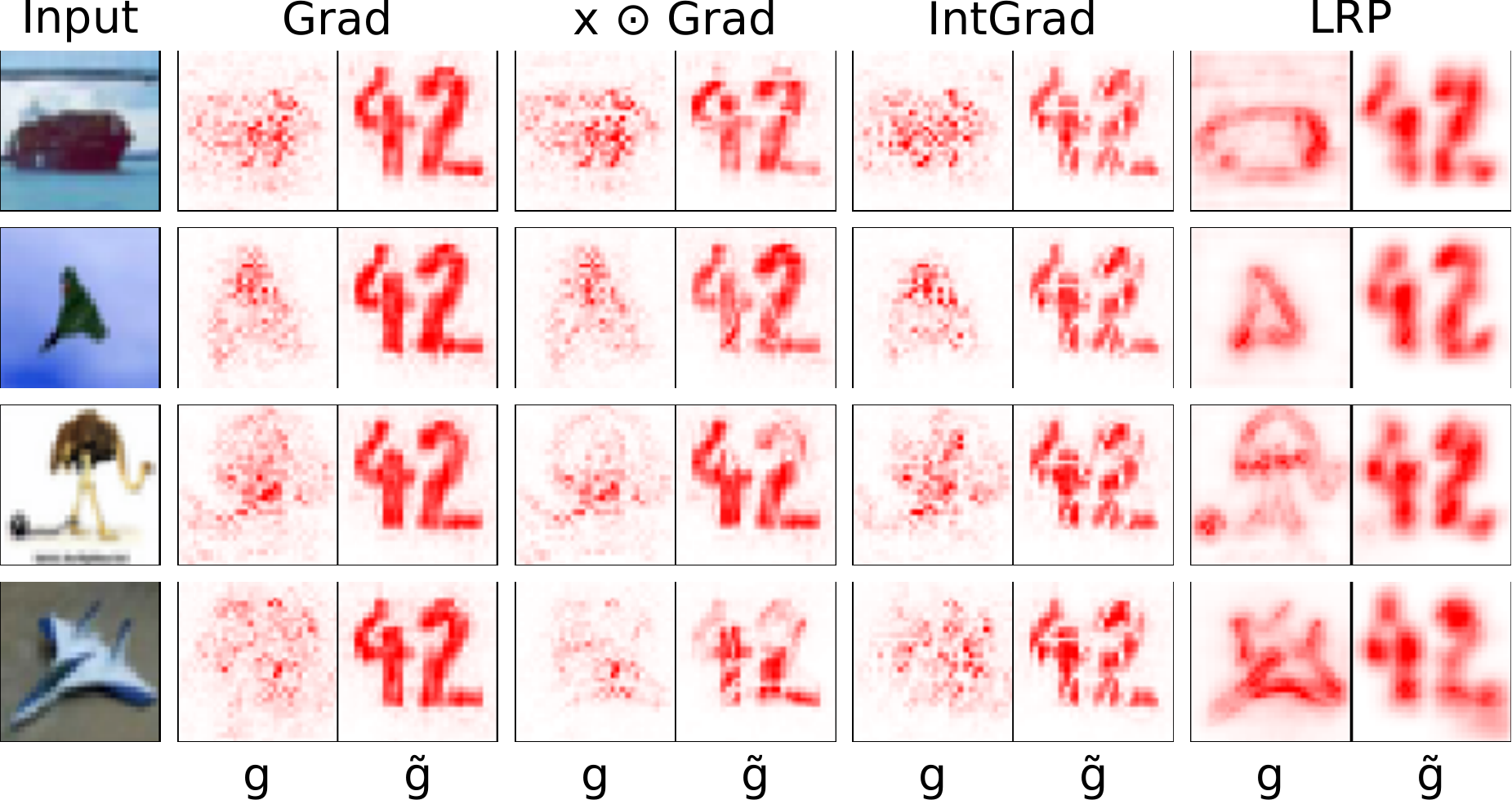}
   \caption{%
   Example explanations from the original model $g$ (left) and the manipulated model $\tilde{g}$ (right). %
   Images from the test sets of FashionMNIST  (\textbf{top}) and CIFAR10 (\textbf{bottom}).%
   } \label{fig:exampleManipulations}
\end{figure}

Figure~\ref{fig:exampleManipulations} illustrates this for examples from the FashionMNIST and CIFAR10 test sets.
We stress that we use a single model for Gradient, x$\odot$Grad, and Integrated Gradient methods which demonstrates that the manipulation generalizes over all considered gradient-based methods. 

The left-hand-side of Figure~\ref{fig:quant} shows quantitatively that manipulated model $\tilde{g}$ closely reproduces the target map $h^t$ over the entire test set of FashionMNIST.
We refer to the Appendix~\ref{app:plots} for additional similarity measures, examples, and quantitative analysis for all datasets.

\section{Robust Explanations}
Having demonstrated both theoretically and experimentally that explanations are highly vulnerable to model manipulation, we will now use our theoretical insights to propose explanation methods which are significantly more robust under such manipulations.

\subsection{TSP Explanations: Theory}
In this section, we will define a robuster \emph{gradient} explanation method.
Appendix~\ref{app:tsp} discusses analogous definitions for other methods.

We can formally define an explanation field $H_g$ which associates to every point $x$ on the data manifold $S$ the corresponding gradient explanation $h_g(x)$ of the classifier $g$.
We note that $H_g$ is generically a vector field \emph{along} the manifold since $h_g(x)\in \mathbb{R}^D \cong T_x M$, i.e. it is an element of the tangent space $T_xM$ of the embedding manifold $M$ and \emph{not} an element of the tangent space $T_xS$ of data manifold $S$.

As explained in Section~\ref{sec:differentialGeom}, we can decompose the tangent space $T_pM$ of the embedding manifold $M$ as follows $T_xM = T_xS \oplus T_xS^\bot$.
Let  $P:T_xM \to T_x S$ be the projection on the first summand of this decomposition.
We stress that the form of the projector $P$ depends on the point $x \in S$ but we do not make this explicit in order to simplify notation.
We can then define:

\begin{definition}
The tangent-space-projected (tsp) explanation field $\hat{H}_g$ is a vector field on the data manifold $S$.
It associates to each $x \in S$, the tangent-space-projected (tsp) explanation $\hat{h}_g(x)$ given by
\begin{align}
    \hat{h}_g(x) = (P \circ h_g) \,(x) \in T_x S \,.
\end{align}
\end{definition}

Intuitively, the tsp-explanation $\hat{h}_g(x)$ is the explanation of the model $g$ projected on the "tangential directions" of the data manifold.

We recall from our discussion of Theorem~\ref{th:explanation} that we can always find classifiers $\tilde{g}$ which coincide with the original classifier $g$ on the data manifold $S$ but may differ in the gradient components orthogonal to the data manifold, i.e. for some $x\in S$ it holds that
\begin{align*}
(1-P) \, \nabla g(x) \neq (1-P) \, \nabla \tilde{g}(x) \,.
\end{align*}
On the other hand, the components tangential to the manifold $S$ agree
\begin{align*}
P \, \nabla g(x) = P \, \nabla \tilde{g}(x) \,,  && \forall x \in S   \,.
\end{align*}
In other words, the tsp-gradient explanations of the original model $g$ and any such model $\tilde{g}$ are identical:
\begin{align}
\hat{h}_g(x) = \hat{h}_{\tilde{g}}(x) && \forall x\in S \,.
\end{align}
It can therefore be expected that tsp-explanations $\hat{h}_g$ are significantly more robust compared to their unprojected counterparts $h_g$.

For other explanation methods, the corresponding tsp-explanations may be obtained using a slightly modified projector $P$.
We refer to Appendix~\ref{app:tsp} for more details.

\subsection{TSP Explanations: Methods}

\begin{figure}
  \centering
  \includegraphics[width=1.\linewidth]{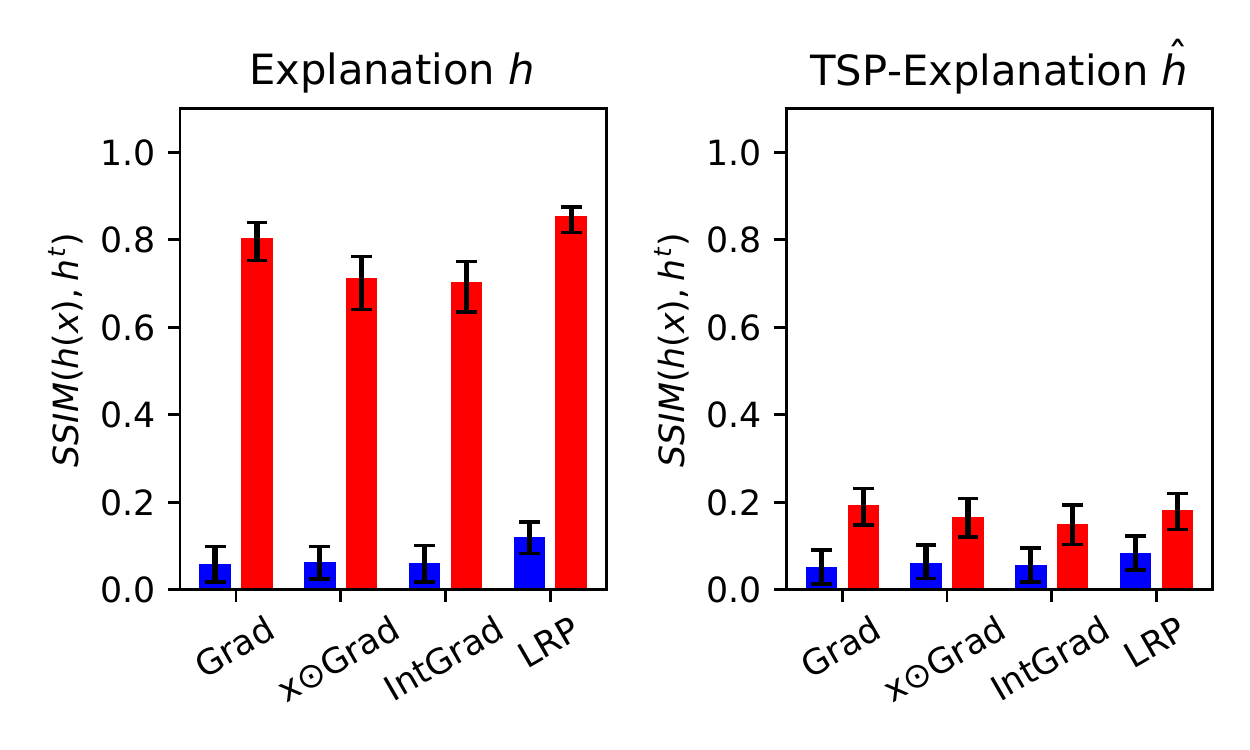}
  \caption{%
  \textbf{Left:} SSIM of the target map $h^t$ and explanations of \original{original model } and \manipulated{manipulated} respectively. %
  Clearly, the manipulated model $\tilde{g}$ has explanations which closely resemble the target map $h^t$ over the entire FashionMNIST test set. %
  \textbf{Right:} Same as on the left but for \emph{tsp-explanations}. %
  The model $\tilde{g}$ was trained to manipulate the tsp-explanation. %
  Evidently, tsp-explanations are considerably more robust than their unprojected counterparts on the left. %
  Colored bars show the median. %
  Errors denote the 25th and 75th percentile. %
  Other similarity measures show similar behaviour and can be found in Appendix~\ref{app:plots}. %
  }
  \label{fig:quant}
\end{figure}

\textbf{Flat Submanifolds and Logistic Regression:}
Recall from Section~\ref{sec:manipulationMethods} that for a logistic regression model $g(x) = \sigma( w^T x + c)$ with gradient explanation $h^{\text{grad}}_g=w$, we can define a manipulated model
$$\tilde{g}(x)=\sigma\left(w^T x +  \sum_i \lambda_i ( \hat{w}^{(i)^T} x - b_i)  + c \right)$$

with gradient explanation $h^{\text{grad}}_{\tilde{g}} = w + \sum_i \lambda_i \hat{w}^{(i)}$ for arbitrary $\lambda_i \in \mathbb{R}$.
Since the vectors $\hat{w}^i$ are normal to the data hypersurface $S$, it holds that $P \hat{w}_i = 0$.
As a result, the gradient tsp-explanations of the original model $g$ and its manipulated counterpart $\tilde{g}$ are identical, i.e.
\begin{align}
\hat{h}^{\textrm{grad}}_{g} = \hat{h}^{\textrm{grad}}_{\tilde{g}} = P w \,.
\end{align}

\begin{figure}[ht!]
  \centering
  \includegraphics[width=1\linewidth]{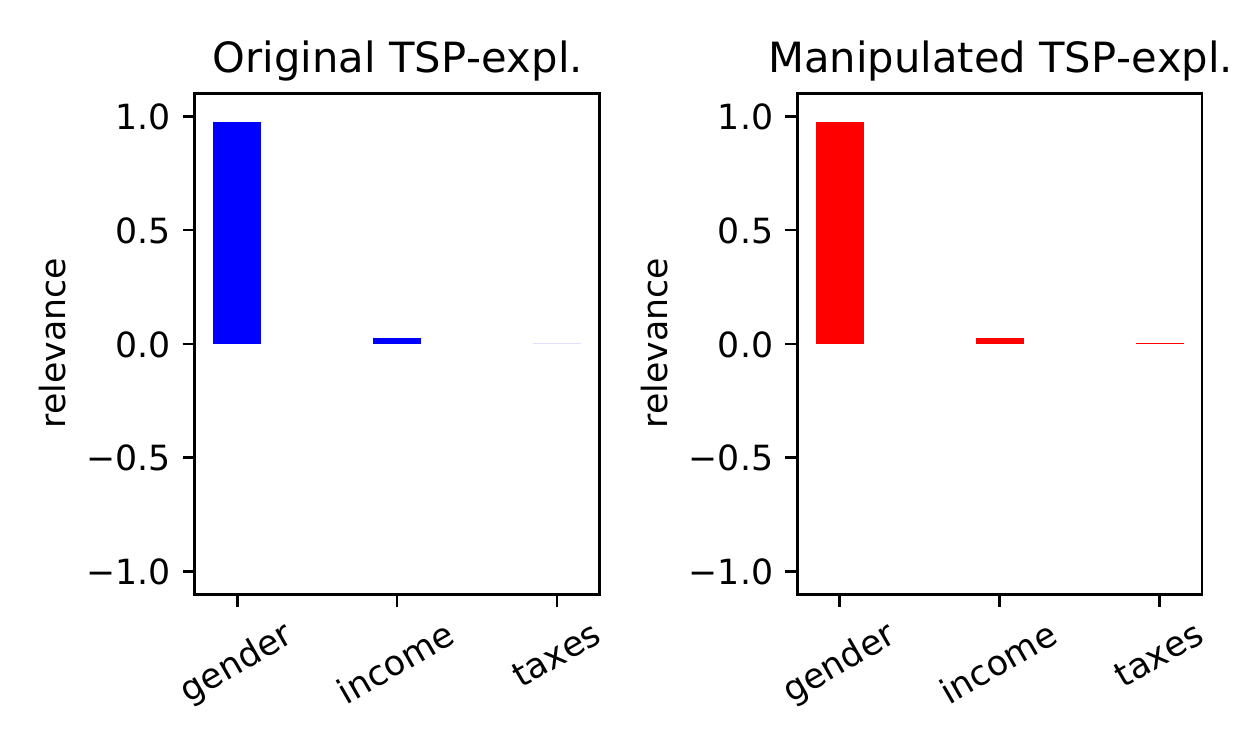}
   \caption{%
       x$\odot$Grad \emph{tsp-explanations} for \original{original classifier} and \manipulated{manipulated} highlight the same features.
       Colored bars show the median of the explanations over multiple examples.%
   }\label{fig:creditxGrad:tsp}
\end{figure}

We discuss the case of other explanation methods in the Appendix~\ref{app:tspFlat}.

\textbf{General Case:} In many practical applications, we do not know the explicit form of the projection matrix $P$.
In these situations, we propose to construct $P$ by one of the following two methods:

\underline{Hyperplane method:} for a given datapoint $x\in S$, we find its $k$-nearest neighbours $x_1, \dots, x_k$ in the training set.
We then estimate the data tangent space $T_xS$ by constructing the $d$-dimensional hyperplane with minimal Euclidean distance to the points $x$, $x_1$, $\dots$, $x_k$.
Let this hyperplane be spanned by an orthonormal basis $q_1, \dots q_d \in \mathbb{R}^D$.
The projection matrix $P$ on this hyperplane is then given by
\begin{align*}
    P = \sum_{i=1}^d q_i \, q_i^T  \,.
\end{align*}

\underline{Autoencoder method:} the hyperplane method requires that the data manifold is sufficiently densely sampled, i.e. the nearest neighbors are small deformations of the data point itself.
In order to estimate tangent space for datasets without this property, we use techniques from the well-established field of manifold learning.
Following \cite{shao2018riemannian}, we train an autoencoder on the dataset and then perform an SVD decomposition of the Jacobian of decoder $D$,
\begin{align}
    \frac{\partial D}{\partial z} = U \, \Sigma \, V \,.
\end{align}
The projector is constructed from the left-singular values $u_1, \dots, u_d \in \mathbb{R}^D$ corresponding to the $d$ largest singular values.
The projector is obtained by
\begin{align}
    P = \sum_{i=1}^d u_i \, u_i^T \,.
\end{align}
The underlying motivation for this procedure is reviewed in Appendix~\ref{app:autoencoderMotivation}.

After one of these methods is used to estimate the projector $P$ for a given $x\in S$, the corresponding tsp-explanation can be easily computed by $\hat{h}(x) = P \, h(x)$.

\subsection{TSP Explanations: Practice}
In this section, we will apply tsp-explanations to the examples of Section~\ref{sec:attackExperiments} and show that they are significantly more robust under model manipulations.

\begin{figure}
  \centering
  \includegraphics[width=1\linewidth]{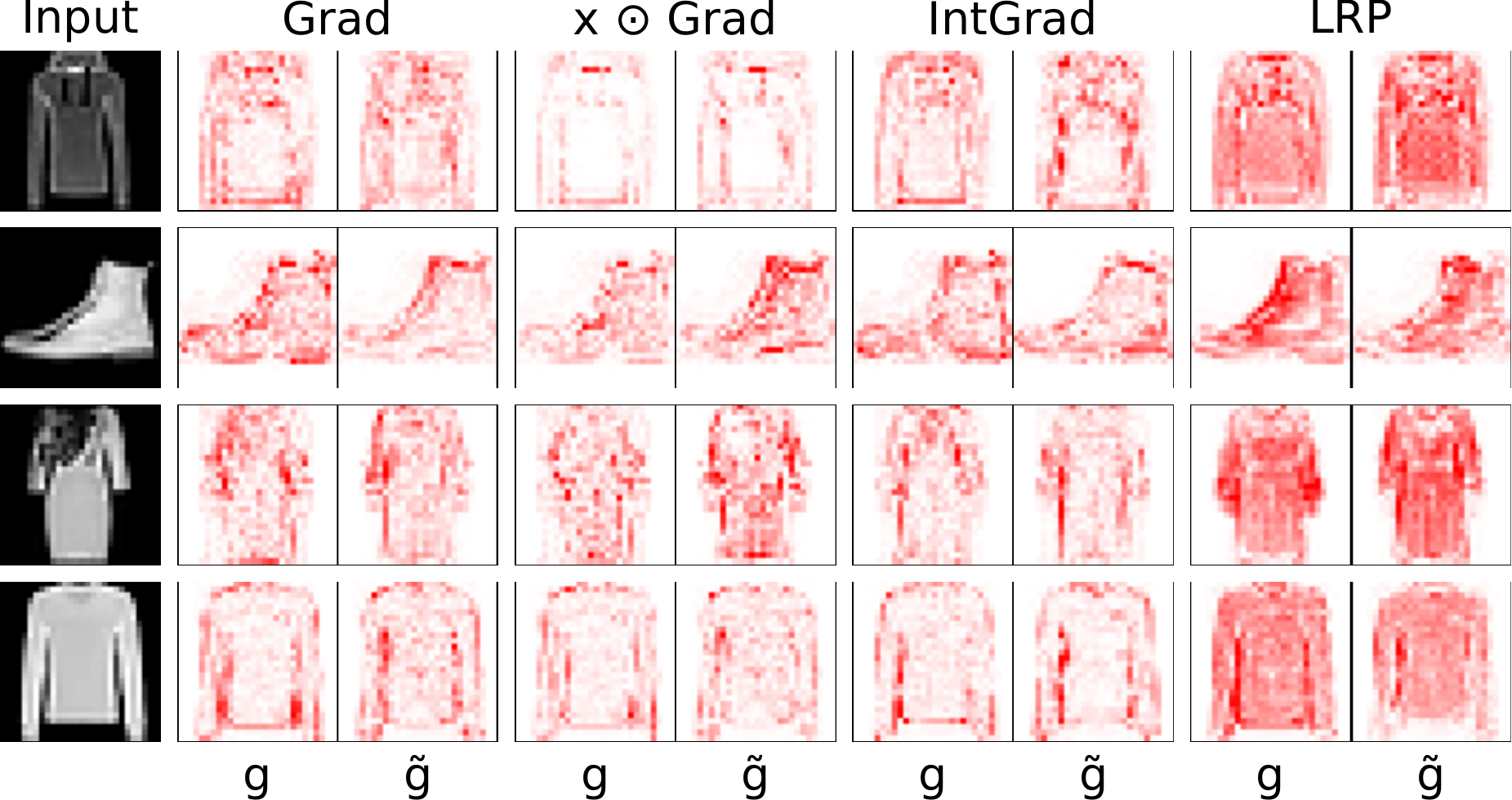}

  \vspace{0.3cm}

  \includegraphics[width=1\linewidth]{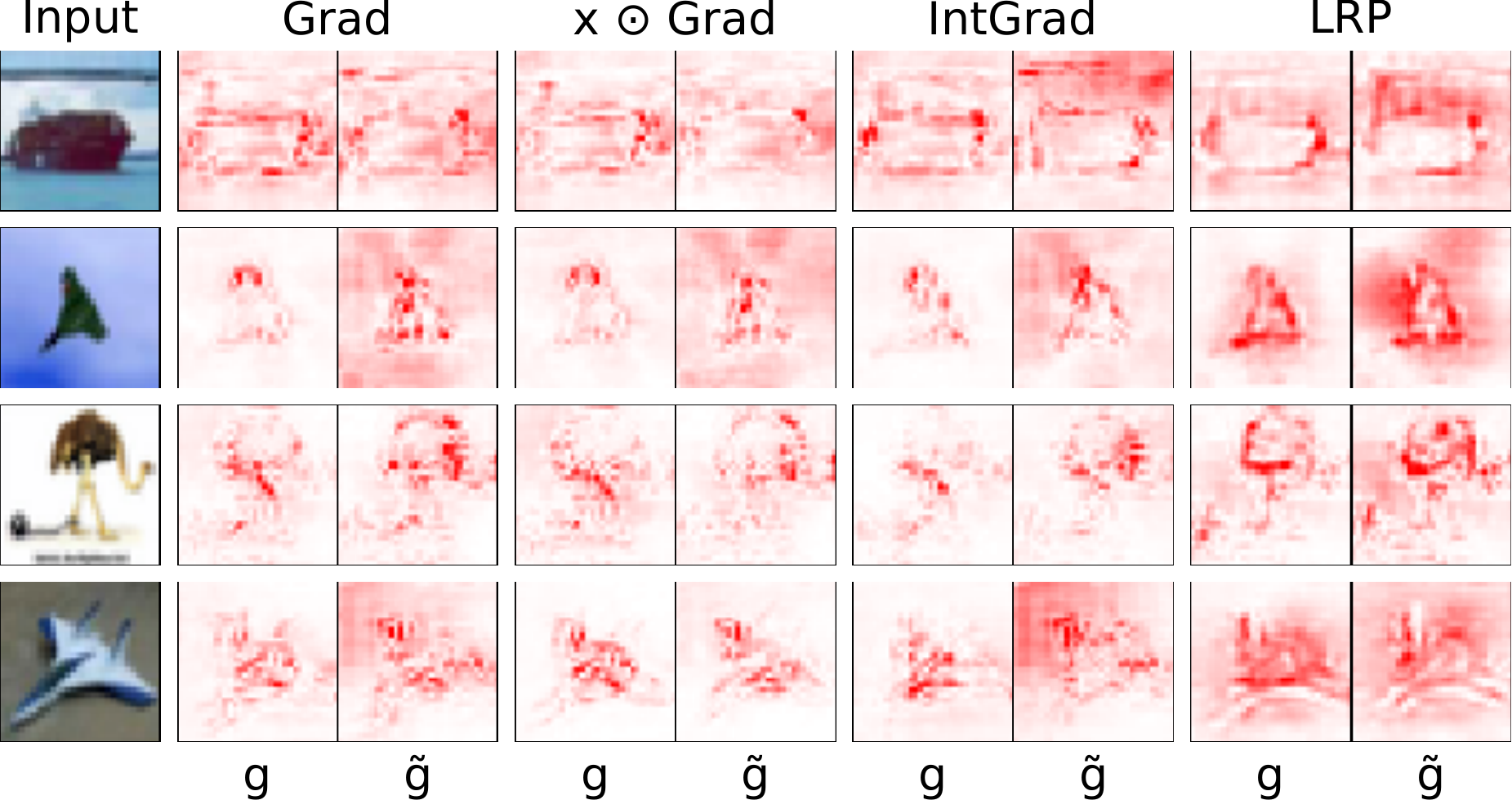}
  \caption{%
Tsp-explanations for the models and images of Figure~\ref{fig:exampleManipulations}. %
The tsp-explanations of the original model $g$ and manipulated $\tilde{g}$ are similar suggesting that the manipulations were mainly due to components orthogonal to the data manifold.%
  }
  \label{fig:projected}
\end{figure}

\paragraph{Credit Assessment:}
From the arguments of the previous section, it follows that the explanations of the manipulated and original model agree.
We indeed confirm this experimentally, see Figure~\ref{fig:creditxGrad:tsp}.
We refer to the Appendix~\ref{app:credit} for more details.

\paragraph{Image Classification:} For MNIST and FashionMNIST, we use the hyperplane method to estimate the tangent space.
For CIFAR10, we find that the manifold is not densely sampled enough and we therefore use the autoencoder method. This is computationally expensive and takes about 48h using four Tesla P100 GPUs. We refer to Appendix~\ref{app:plots} for more details.

Figure~\ref{fig:projected} shows the tsp-explanations for the examples of Figure~\ref{fig:exampleManipulations}.
The explanation maps of the original and manipulated model show a high degree of visual similarity.
This suggests the manipulation occurred mainly in directions orthogonal to the data manifold (as the tsp-explanations are obtained from the original explanations by projecting out the corresponding components).
This is also confirmed quantitatively, see Appendix~\ref{app:plots}.
Furthermore, tsp-explanations tend to be considerably less noisy than their unprojected counterparts (see Figure \ref{fig:projected} vs \ref{fig:exampleManipulations}). This is expected from our theoretical analysis: consider gradient explanations for concreteness. Their components orthogonal to the data manifold are undetermined by training and are therefore essentially chosen at random. This fitting noise is projected out in the tsp-explanation which results in a less noisy explanation. 

If the adversaries knew that tsp-explanations are used, they could also try to train a model $\tilde{g}$ which manipulates the tsp-explanations directly.
However, tsp-explanations are considerable more robust to such manipulations, as shown on the right-hand-side of Figure~\ref{fig:quant}.

We refer to Appendix~\ref{app:plots} for more detailed discussion.

\section{Conclusion}
A central message of this work is that widely-used explanation methods should not be used as proof for a fair and sensible algorithmic decision-making process.
This is because they can be easily manipulated as we have demonstrated both theoretically and experimentally.
We propose modifications to existing explanation methods which make them more robust with respect to such manipulations.
This is achieved by projecting explanations on the tangent space of the data manifold.
This is exciting because it connects explainability to the field of manifold learning.
For applying these methods, it is however necessary to estimate the tangent space of the data manifold.
For high-dimensional datasets, such as ImageNet, this is an expensive and challenging task. Future work will try to overcome this hurdle. Another promising direction for further research is to apply the methods developed in this work to other application domains such as natural language processing.

\section*{Acknowledgements}
We thank the reviewers for their valuable feedback.
P.K. is greatly indebted to his mother-in-law as she took care of his sick son and wife during the final week before submission.
We acknowledge Shinichi Nakajima for stimulating discussion.
K-R.M. was supported in part by the German Ministry for Education and Research (BMBF) under Grants 01IS14013A-E, 01GQ1115, 01GQ0850, 01IS18025A and 01IS18037A.
This work is also supported by the Information \& Communications Technology Planning \& Evaluation (IITP) grant funded by the Korea government (No. 2017-0-001779), as well as by the Research Training Group "Differential Equation- and Data-driven Models in Life Sciences and Fluid Dynamics (DAEDALUS)" (GRK 2433) and Grant Math+, EXC 2046/1, Project ID 390685689 both funded by the German Research Foundation (DFG).

\bibliography{refs}
\bibliographystyle{icml2020}

\clearpage
\appendix
\section{Proofs}
\subsection{Theorem~\ref{th:extension}}\label{app:extensionTheoremProof}

We first recall a few basic definitions and theorems necessary for the proof of Theorem~\ref{th:extension}. Our presentation will be necessarily brief as it can hardly replace a course on differential geometry. However we closely follow \cite{smoothmanifold} to which we refer for a more detailed and complete presentation.

\begin{definition}
An embedded submanifold $S$ of $M$ is a subset $S \subset M$ that is itself a manifold (with respect to the subspace topology) endowed with a smooth structure with respect to which the inclusion map $S \xhookrightarrow{} M$ is a smooth embedding.
If $S$ is closed as a set, the submanifold is called properly embedded.
\end{definition}

Let $U$ be an open subset of $\mathbb{R}^n$ and $k \in \{1, \dots, n \}$. A \emph{k-slice} of $U$ is any subset $S \subset U$ of the form
\begin{equation*}
    S = \{ (x^1, \dots, x^k, 0, \dots 0 ) \in U \} \,.
\end{equation*}
We say that a submanifold $S \subset M$ satisfies the local k-slice condition if each point $p \in S$ is contained in the domain of a chart $(U, \phi)$ for which $\phi(S \cap U)$ is a single k-slice in $\phi(U)$.

\begin{theorem}
An embedded $k$-dimensional submanifold $S$ satisfies the local $k$-slice condition.
\end{theorem}
We refer to Theorem~5.8 of \cite{smoothmanifold} for a proof.

\begin{definition}
Let M be a smooth manifold and $S \subset M$ an embedded submanifold. A vector field $X$ along $S$ assigns to each $p \in S$ a vector $X_p \in T_p M$.
\end{definition}

For each $p \in S$, we can decompose the tangent space $T_p M = T_p S \oplus T_p S^\bot$, where $T_p S^\bot$ is the orthogonal complement of $T_p S$.

A standard tool for extending functions from a local coordinate patch to the entire manifold is given by the following definition:
\begin{definition}
Let M be a topological space and $\Phi=(\phi_\alpha)_{\alpha \in I}$ an open cover indexed by the set $I$. A partition of the unity subordinate to $\Phi$ is an family $(\psi_\alpha)_{\alpha \in I}$ of continuous functions $\psi_\alpha: M \to \mathbb{R}$ with the properties:
\begin{enumerate}
    \item $\forall x \in M$ and $\forall \alpha \in I$: $0 \le \psi_\alpha(x) \le 1$
    \item $\forall \alpha \in I$: $\text{supp} (\psi_\alpha) \subset \phi_\alpha$
    \item $(\text{supp} \,\psi_\alpha)_{\alpha \in I}$ is locally finite, i.e. $\forall p \in M$, $\exists U \subset M$ such that $U \cap \text{supp}(\psi_\alpha)\neq\emptyset$ for only finitely many values of $\alpha$.
\end{enumerate}
\end{definition}
It can be shown that for any open cover of a manifold $M$, a partition of the unity subordinate to this cover exists. We refer to Theorem~2.23 of \cite{smoothmanifold} for a proof.

Our main theorem is a generalization of the well-known submanifold extension lemma (see, for example, Lemma~5.34 in \cite{smoothmanifold}). While we could not find such a generalization in the literature, we suspect that it is entirely obvious to differential geometers but typically not needed for their purposes. We now state this main theorem before giving a proof:

\begin{theorem}
Let $S \subset$ M be a properly embedded $d$-dimensional submanifold of the $D$-dimensional manifold $M$ and $V = \sum_{i=d+1}^{D} v^i \partial_i$ a smooth vector field along $S$ which for each $p\in S$ assigns vectors in $T_p M^\bot$. For any smooth function $f:S \to \mathbb{R}$, there exists a smooth extension $F: M \to \mathbb{R}$ such that $F|_{S}=f$ and
\begin{align*}
    \nabla F(x) = ( \nabla_1 f(x), \dots \nabla_d f(x), v^1(x), \dots, v^{D-d}(x))
\end{align*}
for $x \in S$.
\end{theorem}

\textbf{Proof:} Since $S$ is embedded, there exists a slice chart $(U_p, \phi_p)$ for each $p \in S$. We extend $f$ in $U_p$ by the smooth map
\begin{align*}
    F_p (x_1, \dots, x_D) = f(x_1, \dots, x_d) + \sum_{I=d+1}^D v^I (x_1, \dots, x_d) \, x^I \,.
\end{align*}
By the definition of a slice chart, $\phi(p)=(x_1, \dots, x_d, 0, \dots, 0)$ for $p\in S$. Therefore, it follows that
\begin{align*}
    F |_{S} = f \,.
\end{align*}
Let $\{\psi_p, p \in S\} \cup \{ \chi \} $ be a partition of unity subordinate to the open cover $\{U_p; \, p\in S\} \cup \{M  \setminus S \}$.\footnote{We note that $M  \setminus S$ is open since $S$ is closed.} We define
\begin{align*}
    F(x)=\sum_{p \in S} \psi_p(x) F_p(x) \,.
\end{align*}
For $x\in S$, it holds that $F_p(x)=f(x)$ and thus $F(x)=f(x)\sum_{p \in S} \psi_p(x) = f(x)$ because $\sum_{p \in S} \psi_p(x)=1$. Since the collection of supports of the $\psi_p$ is locally finite, $F$ is smooth.

The gradient of $F$ at $x\in S$ can be straightforwardly calculated. For $I\in\{d+1,\dots,D\}$, one obtains
\begin{align*}
    \nabla_I F(x) &= \nabla_I \sum_p \psi_p(x) F_p(x) \\
    &= \sum_p \nabla_I \psi_p(x) \, F_p(x) + \sum_p \psi_p(x) \nabla_I F_p(x) \\
    &= f(x) \, \nabla_I \sum_p \psi_p(x) + \sum_p \psi_p(x) v^I(x) \,.
\end{align*}
We note that sum and differentiation commute due to the local finiteness of the partition $\psi$. Using $\sum_p \psi_p(x)=1$, it follows that $\partial_I \sum_p \psi_p(x)=0$. We thus have derived that
\begin{equation*}
 \partial_I F(x)=v^I(x) \sum_p \psi_p(x)= v^I(x) \,.
\end{equation*}
For $i\in\{1,\dots,d\}$, one obtains
\begin{align*}
    &\nabla_i \sum_p \psi_p(x) F_p(x) \\
    =&  \sum_p  \nabla_i\psi_p(x) \, F_p(x) + \sum_p \psi_p(x) \, \nabla_i F_p(x)  \\
    =& f(x) \, \nabla_i \sum_p \psi_p(x) + \\& \;\;\;\; + \sum_p \psi_p(x) \left( \nabla_i f(x) + \sum_{I=d+1}^D  x^I \; \nabla_i v^I(x)  \right) \,.
\end{align*}
The first term vanishes due to $\nabla_i \sum_p \psi_p(x) = \nabla_i 1 = 0$. For the last term, we use that for $x\in S$ it holds that $x^I=0$. As a result, we derive that
\begin{align*}
    \nabla_i F(x) = \nabla_i f(x) \,.
\end{align*}
$\square$

\subsection{Theorem~\ref{th:explanation}}\label{app:proofOtherMethods}
\subsubsection{Bounds on Explanations}\label{app:boundExpl}
As noted in the main text, a global rescaling of the explanation maps $h$ is merely conventional. A natural convention is to bound the explanations such that $h_i \in [-0.5, 0.5]$ for all $i=1\dots D$. For the gradient map, this can be ensure by defining $h(x) = \lambda \, \nabla g(x)$ where $\lambda = \tfrac{1}{C}$ (since by assumption $|\nabla_i g(x)| \le C$). In particular, all target explanation maps are then chosen to obey this bound. 
For convenience, we can absorb rescaling $\lambda$ in the classifier $g$ by redefining $g \to \lambda g$. As a result, we always choose the convention that $|\nabla_i g(x)|\le 1$ without loss of generality.

More generally, let $h$ denote any bounded explanation method
\begin{align}
    |h_i(x)|\le C \in \mathbb{R}_+ && \forall x \in S \label{eq:appBoundOnExpl}
\end{align}
We note that all considered explanation maps obey
\begin{align}
g \to \lambda g &&\Rightarrow && h_g \to \lambda h_g
\end{align}
for $\lambda \in \mathbb{R}$ since they are linear in $g$.

From this, it follows that any bounded explanation method can be assumed to be bounded by $0.5$ because this can be ensured by an irrelevant rescaling. We again adopt the convention in which this rescaling factor is absorbed in $g$.

\subsubsection{Proofs for other Explanations}
In this appendix, we will proof Theorem~\ref{th:explanation} for $x\odot \text{Grad}$ and $\epsilon$-LRP.

\textbf{$\mathbf{x}\odot \textrm{Grad}$}: We assume that the explanation map of $g$ is bounded, i.e. $|h^g_i(x)|=|(x \odot \nabla_i g(x))_i|\le C \in \mathbb{R}_+$ for all $x\in S$. We furthermore assume that there exists a chart for which the coordinates $x_i \neq 0$ are non-vanishing for $i>d$. In practice, this can be easily ensured by an appropriate shift of the data.\footnote{If we do not allow for the freedom of shifting the data, any valid $\mathbf{x}\odot \text{Grad}$ explanation map must have zero relevance for input components $x_i$ which are vanishing. If one restrict the target map $h^t$ to be valid, no shifts are needed for the proof.} Given a target explanation $h^t(x)$, we choose a extension $G$ of $g|_S$ such that
\begin{align*}
    \nabla G(x) = ( \nabla_1 g(x), \dots \nabla_d g(x), \tfrac{h^t_{d+1}(x)}{x_{d+1}}, \dots, \tfrac{h^t_{D}(x)}{x_D}) \,.
\end{align*}
The explanation of $G$ is given by $h_G(x)=x \odot \nabla G(x)$. The mean-squared error between target and model explanation is then given by
\begin{align*}
\text{MSE}(h_G(x), h^t(x))
= \tfrac1D \sum_{i=1}^D (x_i \nabla_i G(x) - h^t_i(x))^2
\end{align*}
This sum can be decomposed as
\begin{align*}
&\tfrac1D \sum_{i=1}^d (x_i \nabla_i g(x) - h^t)^2 + \tfrac1D \sum_{i=d+1}^D x_i^2 (\nabla_i G(x) - \tfrac{h^t_i(x)}{x_i})^2
\end{align*}
Using the fact that we can assume $|h^{g}_i|=|x_i \nabla_i g(x)|\le 0.5$ without loss of generality\footnote{We note that the necessary rescaling of $h^g$ is not in conflict with the shift to ensure $x_i\neq 0$ because the latter condition is scale-invariant.} and that we can rescale $h^t$ arbitrarily, it then follows
\begin{align*}
\text{MSE}(h_G(x), h^t(x) & \le \frac{d}{D} \,.  
\end{align*}

\paragraph{$\mathbf{\epsilon}$-LRP:} We assume that the network uses $\relu$ non-linearities. In fact, LRP can be shown to be theoretically well-motivated under this assumption by using Deep Taylor Decomposition \cite{dtd}.

It can be shown that $\epsilon$-LRP can be mathematically reformulated as
\begin{align*}
    h_{\epsilon\text{LRP}} = x \odot \tilde{\nabla} g(x) \,,
\end{align*}
where the operator $\tilde{\nabla}$ acts on non-linearities $f$ by
\begin{align}
\tilde{\nabla} f(z)=\tfrac{f(z)}{z}
\end{align}
and on affine linear functions as the standard gradient $\nabla$. We refer to the Appendix~A of \cite{deeplift} for a proof.
By our assumption, all non-linearities are $\relu$ and therefore obey
$$\tilde{\nabla}\relu(x)=\theta(x) \,$$
where $\theta(x)$ is the Heaviside step function. This coincides with normal gradient operator $\nabla \relu(x) = \theta(x)$. This observation was, to the best of our knowledge, first made in \cite{deeplift}. Therefore, the proof for $x\odot\text{Grad}$ applies verbatim for this method as well. $\square$

\subsection{Flat Manifolds and other Explanation Methods}\label{app:flatOtherMethods}
It was shown in the main text that one can always construct a model

\begin{align}
    \tilde{g}(x) = \sigma\left(w^T x +  \sum_i \lambda_i ( \hat{w}^{(i)^T} x - b_i)  + c \right) \,,
\end{align}

which agrees with $g(x) = \sigma (w^T x + c)$ for all datapoints $x \in S$ but has gradient explanation map
\begin{align}
    h_{\text{grad}}(x) = w + \sum_i \lambda_i \hat{w}^{(i)} \,.
\end{align}
By choosing $\lambda_i$ appropriately, we can always set components of $h_{\text{grad}}$ corresponding to orthogonal directions $\hat{w}_i$ of the data $S$ to an arbitrary $h^t_i$, i.e.
$$
\lambda_i = h^t_i - w^T \hat{w}^{(i)}
$$
where we have normalized $\hat{w}^{(i)}$ such that it has unit norm. For $x \odot \text{Grad}$, we can similarly choose
$$
\lambda_i = \frac{h^t_i - (x \odot w)^T\hat{w}^{(i)}}{(x \odot \hat{w}^{(i)})^T \hat{w}^{(i)}}
$$
As already discussed in Appendix~\ref{app:proofOtherMethods}, valid $x \odot \text{Grad}$ explanations map have to be zero in components $h_i$ for which the corresponding input component $x_i$ are vanishing. As a result, one only needs to set $\lambda_i$ to a non-vanishing value if $x_i \neq 0$. Thus, the expression above is well-defined for all valid explanation maps. The corresponding statement for $\epsilon$-LRP method can be proven completely analogously.

We also note that $\epsilon$-LRP and IntGrad coincide with the x$\odot$Grad method for logistic regression. For the latter, one has to choose a vanishing baseline point $\bar{x}$. The generalization to non-vanishing baselines is however straightforward by substituting $x \to x-\bar{x}$.

\section{Credit Risk using other Explanation Methods}\label{app:credit}
We originally tested our procedure on two credit-risk datasets. Unfortunately, we realized that the licences of these datasets do not permit publication of these results. Since our results only mildly depend on the data (for example, the gradient explanation is completely independent of it), we decided to generate a synthetic dataset as follows: 
the feature 'gender' is sampled with equal probability for the values $1$ for male or $-1$ for female.
The feature 'income' is sampled from a normal distribution with mean $\mu=5000$ and standard deviation $\sigma=5000$. We clipped to a minimum of $250$ to ensure only positive income. We then normalized the income to take values between $0$ and $1$ by dividing by the maximum income. The feature 'taxes' is $0.4 x_{\text{income}}$ and, for simplicity, not further normalized. We use $\lambda = 1000$ as scaling factor for the weights $\hat{w}$ of the modified classifier $\tilde{g}$.

\begin{figure}[ht!]
  \centering
  \includegraphics[width=1\linewidth]{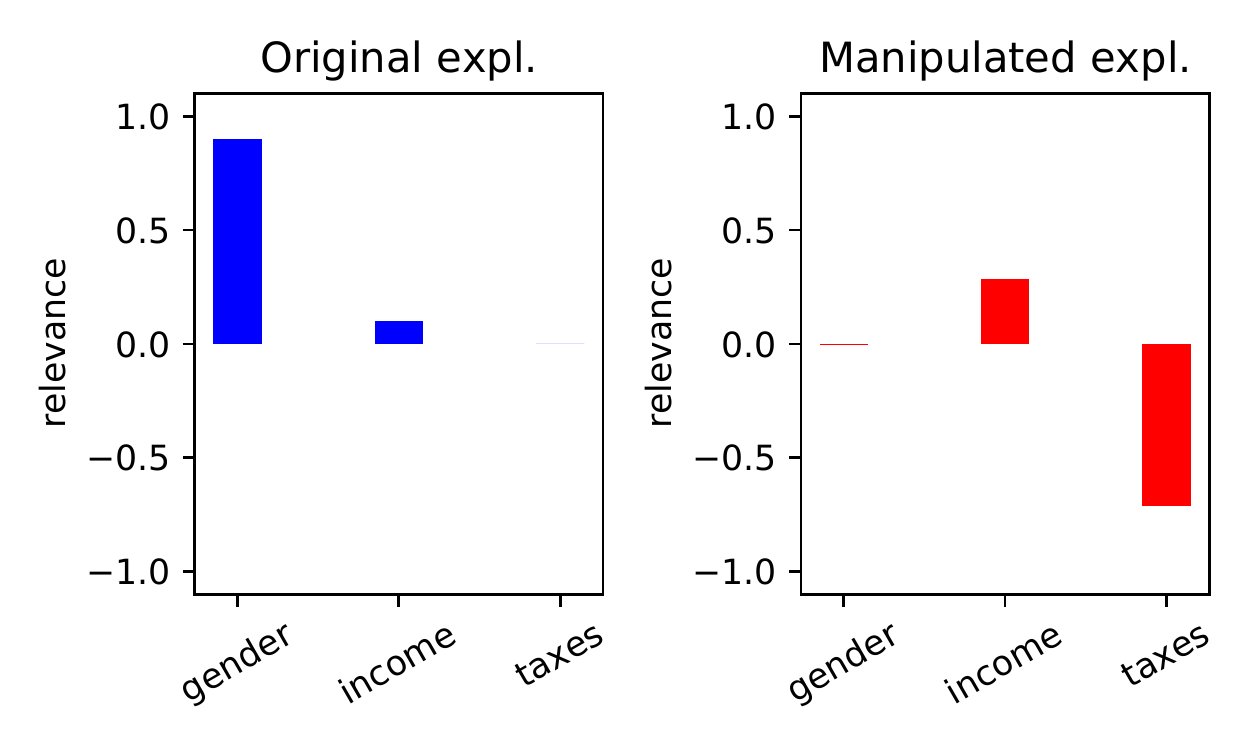}
   \caption{Gradient explanations for classifier $g$ and fairwashed classifier $\tilde{g}$ highlight completely different features.\label{fig:heatmaps_credit_grad}}
\end{figure}

\begin{figure}[ht!]
  \centering
  \includegraphics[width=1\linewidth]{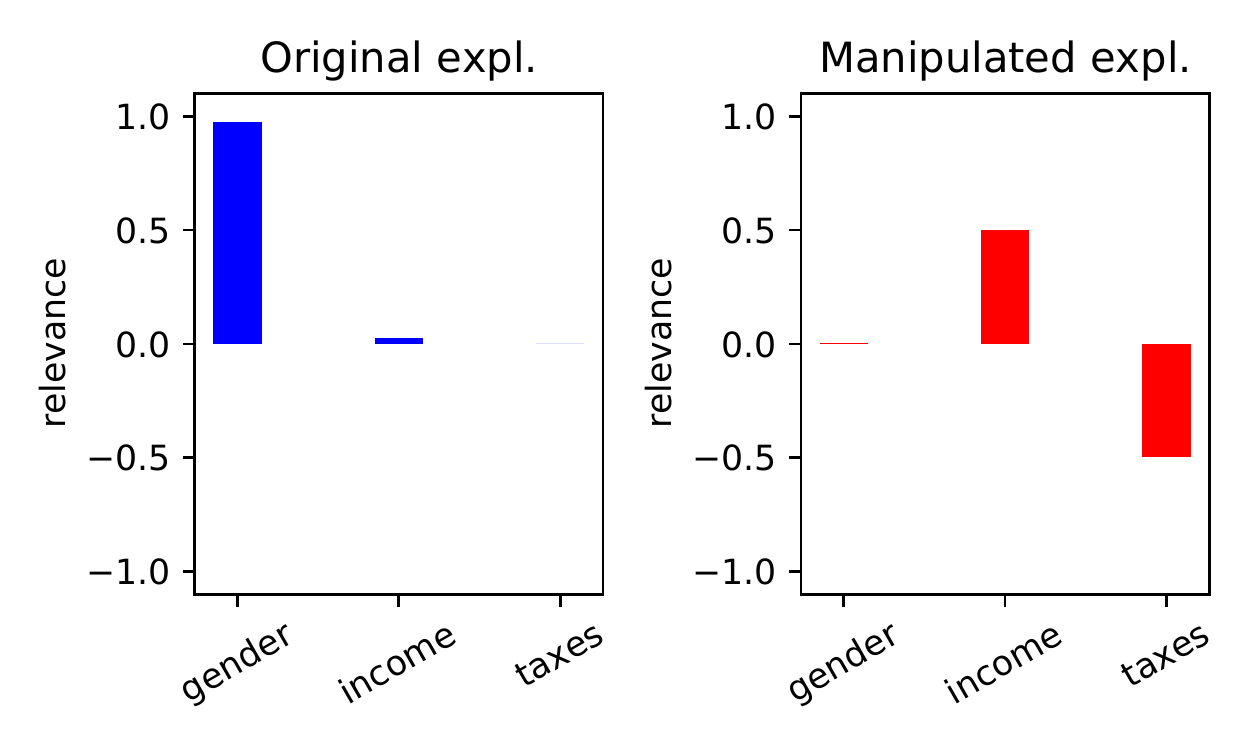}
   \caption{x$\odot$Grad explanations for classifier $g$ and fairwashed classifier $\tilde{g}$ highlight completely different features.\label{fig:heatmaps_credit_grad_x_input}}
\end{figure}

The bars in Figures~\ref{fig:heatmaps_credit_grad} and~\ref{fig:heatmaps_credit_grad_x_input} show the average explanation map with error bars as standard deviations. We only show explanation maps for positive classification results (examples where credit was given). All explanation maps are normalized to have $\sum_i |h_i|=1$.

\begin{figure}[ht!]
  \centering
  \includegraphics[width=1\linewidth]{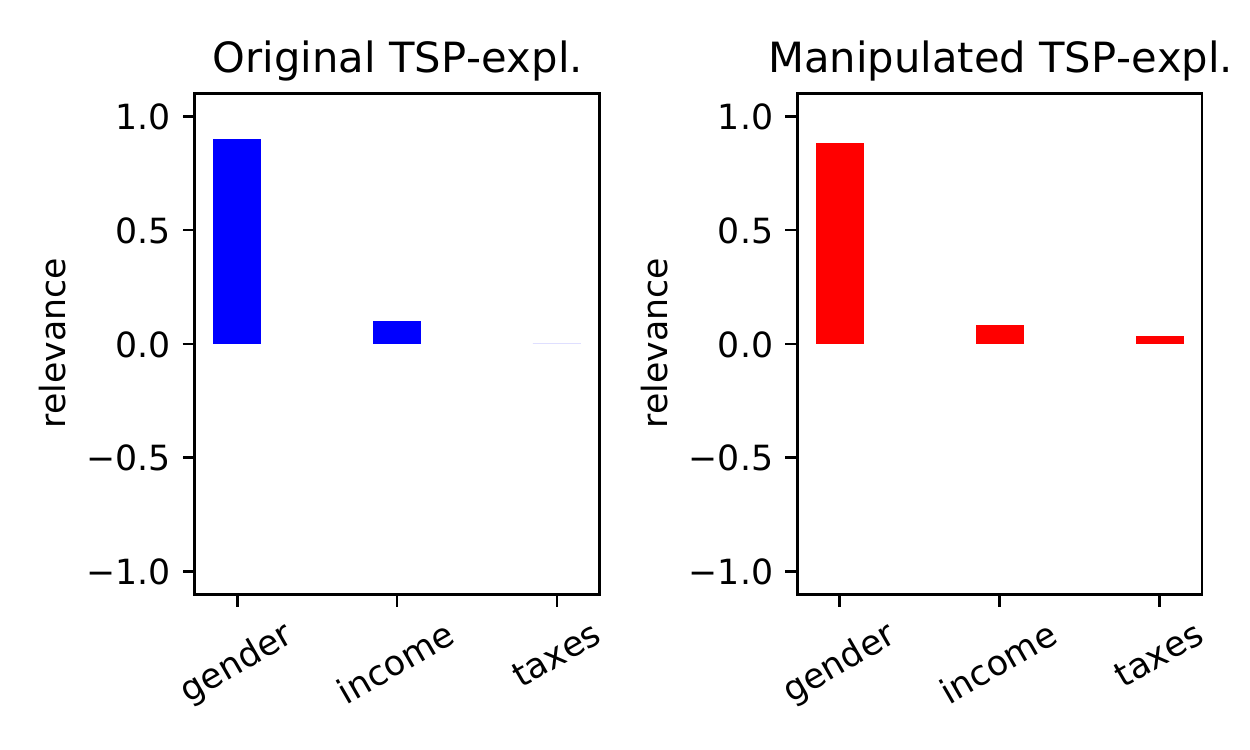}
   \caption{Gradient \emph{tsp-explanations} for \original{original classifier} and \manipulated{manipulated} highlight the same features.%
   Colored bars show the median of the explanations over multiple examples.%
   }
   \label{fig:heatmaps_credit_grad_tsp}
\end{figure}

\begin{figure}[ht!]
  \centering
  \includegraphics[width=1\linewidth]{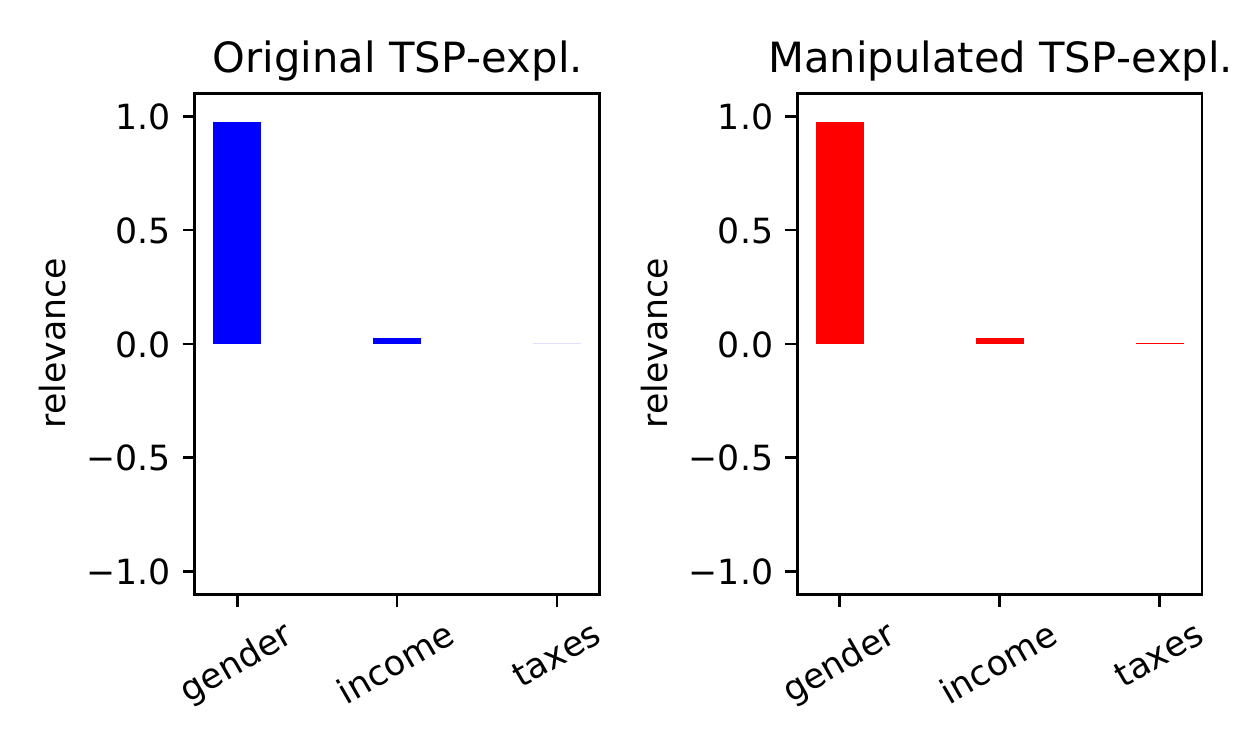}
   \caption{x$\odot$Grad \emph{tsp-explanations} for \original{original classifier} and \manipulated{manipulated} highlight the same features.
   Colored bars show the median of the explanations over multiple examples.%
   }
   \label{fig:heatmaps_credit_grad_x_input_tsp}
\end{figure}

\section{TSP-Explanations} \label{app:tsp}
For the $x \odot \textrm{Grad}$ method, we let the projection operator act only on the gradient factor of the explanation map, i.e.
\begin{equation}
\hat{h}_{\text{xGrad}}(x) = x \odot P \, \nabla g(x) \,. \label{eq:tspxGrad}
\end{equation}
This is equivalent to redefining the projection matrix to
\begin{align}
P_{ij} \to
\begin{cases}
\frac{x_i}{x_j} P_{ij} & \textrm{for} \; x_j \neq 0 \,,\\
0 & \textrm{for} \; x_j = 0 \,. \label{eq:genProj}
\end{cases}
\end{align}
and applying this redefined projection operator on the unprojected map $h_{\text{xGrad}}$, i.e.
\begin{align}
\hat{h}_{\text{xGrad}}(x) = P \, h_{\text{xGrad}}(x) \,.
\end{align}
Analogously, we define for the IntGrad method
\begin{align}
\hat{h}_{\textrm{IntGrad}}(x) = &(x - \bar{x}) \nonumber \\
&\odot \tfrac{1}{N} \, \sum_{k=0}^N P \, \nabla g\left(\bar{x} + \tfrac{k}{N} (x-\bar{x}) \right) \,, \label{eq:tspIntGrad}
\end{align}
where $P$ projects on the tangent space of the point at which the corresponding gradient is calculated. In practice however, we cannot guarantee that all the corresponding points lie on the data manifold $S$. We therefore propose to use the projection operator for the data point $x$ instead. We find empirically that this leads to robuster explanations. This definition can again be reformulated in terms of a redefinition of the projection operator in complete analogy to the case of $x\odot\text{Grad}$.

For the LRP method, we propose to use the generalized projection matrix \eqref{eq:genProj} since $\epsilon$-LRP is equivalent to $x\odot\textrm{Grad}$ for $\relu$ activations (see Appendix~\ref{app:proofOtherMethods}) but we also find empirically that the standard projection matrix on the data manifold leads to more robust explanations.

\subsection{Flat manifold and Logistic Regression}\label{app:tspFlat}
For $x\odot\textrm{Grad}$ method, we again straightforwardly see that the tsp-explanations for $g$ and $\tilde{g}$ agree by applying the definition \eqref{eq:tspxGrad}, i.e.
\begin{align}
\hat{h}_{g} (x) = x \odot P \, \nabla g(x) = x \odot P \, \nabla \tilde{g}(x) = \hat{h}_{\tilde{g}}(x) \,.
\end{align}
The corresponding statement for $\epsilon$-LRP can be proven analogously. The same is true for IntGrad if one assumes that all intermediate point as well as the baseline point are on the data manifold.

\subsection{Autoencoder Method}\label{app:autoencoderMotivation}
In the following, we will first show how the proposed procedure for estimating tangent space arises from certain asymptotic limit of autoencoders. 
\begin{definition}
An asymptotically-trained autoencoder with encoder $E:M \to Z$ and $D: Z \to M$ has zero reconstruction error, i.e.
\begin{align*}
    E_{\text{rc}} = \int_S \textrm{d}^D x \, p_{\textrm{data}}(x) \, || (D\circ E)(x) - x ||^2 = 0 \,,
\end{align*}
where $p_{\textrm{data}}$ is a continuous probability density describing the data. Furthermore, the decoder maps on the data manifold $S$, i.e. 
\begin{align*}
\forall z \in Z: && D(z)\in S \,.
\end{align*}
\end{definition}
The latter condition arises from the fact that we want the decoder to generate data samples from latent representations. We note there is good theoretical and experimental evidence that these conditions hold asymptotically for (at least some of the) popular autoencoder architectures, in particular Variational Autoencoders \citeapp{kingma2014auto}.

\begin{theorem}
For a continuous data distribution $p_{\textrm{data}}$, it holds that 
\begin{align}
    E_{\textrm{rc}} = 0 && \Rightarrow && \forall x \in S: \;\; x = (D\circ E) (x) \,,
\end{align}
i.e. every datapoint $x$ is perfectly reconstructed.
\end{theorem}
\textbf{Proof:} Suppose, there exists a $x_0 \in S$ such that $x_0 \not =(D\circ E) (x_0)$. Since the integrand of $E_{\textrm{rc}}$ is continuous, we can always find an $\epsilon>0$ such that this condition holds for every $x \in [x-\epsilon, x+\epsilon]$. Let $\Delta\in \mathbb{R}_+$ denote the infimum of the integrand on this interval. By positivity of the integrand, it holds that $E_{\textrm{rc}} \ge 2\epsilon\, \Delta >0$. $\square$

This theorem then immediately implies that:
\begin{theorem}
The decoder $D: Z \to S$ of an asymptotically-trained autoencoder is surjective on the data manifold $S$.
\end{theorem}
\textbf{Proof:} Assume the contrary, then there exists a $x\in S$ such that $\not \exists z \in Z$: $D(z)=x$. But by the previous theorem, it has to hold that $z=E(x)$ obeys $D(z)=x$ since the autoencoder has vanishing reconstruction error.$\square$

The differential $\textrm{d}_z D(z)=\frac{\partial D}{\partial z}(z)$ is a linear map from the tangent space of $Z$ to the tangent space of $S$, i.e. $\textrm{d}_z D(z):T_zZ \to T_{D(z)}S$. Since the decoder is surjective, the rank of $\textrm{d}_z D$ is the same as the dimensionality of the data manifold $S$, i.e. $\textrm{rk} (d_z D) = d$. These are basic facts of differential geometry and we refer to Chapter~5 and 6 of \cite{smoothmanifold} for a detailed discussion. As a result, the left-singular vectors $u_1, \dots u_d \in \mathbb{R}^D$, corresponding to the $d$ non-vanishing singular values of the decomposition $d_z D(z) = U \, \Sigma \, V$, span the data tangent space $T_{D(z)} S$.

In the non-asymptotic limit, it cannot be expected that this relation holds exactly. For a sufficiently well-trained autoencoder, it is however reasonable to expect that the left-singular values $u_1,\dots,u_d \in \mathbb{R}^D$ corresponding to the $d$ largest singular values are a good approximation for the basis of the data tangent space.

We stress however that we do not have a rigorous proof for this outside of the asymptotic regime discussed above. We furthermore want to remark that our thinking was heavily inspired by the discussion in \cite{shao2018riemannian} which uses very similar techniques. Last but not least, there are a number of alternative approaches in the literature to estimate tangent space. Notable examples include Contractive Autoencoders \citeapp{rifai2011manifold} and semi-supervised GANs \citeapp{kumar2017improved}. It would be interesting to compare these approaches to the one taken in this paper but we leave this to future work.

\section{Details on Experiments}\label{app:plots}
\paragraph{Model Architecture:}
For FashionMNIST and MNIST, we used a convolutional network with two groups of convolution with 20 and 50 filters of size $5 \times 5$ respectively, relu activation and max-pooling over $2 \times 2$, followed by a dense layer with $500$ outputs, a relu activation, and finally another dense layer with outputs down to the number of classes ($10$).
We used VGG16 \cite{simonyan2015very} for experiments on CIFAR10.

\paragraph{Model Training:}
All images were normalized to mean $0$ and standard deviation $1$ within the training set over all pixels.
For CIFAR10 training, we padded all images with 4 pixels of each side in every dimension, and then randomly cropped back to the original size of $32\times 32$.

The original models for FashionMNIST and MNIST were trained from scratch using standard SGD with a learning rate of $0.01$ and a momentum of $0.5$.
The original VGG-16 model for CIFAR10 was trained also trained using standard SGD, but with a learning rate of $0.05$, momentum of $0.9$ and weight decay of $5\times 10^{-4}$.

\begin{figure}
  \centering
  \includegraphics[width=1.\linewidth]{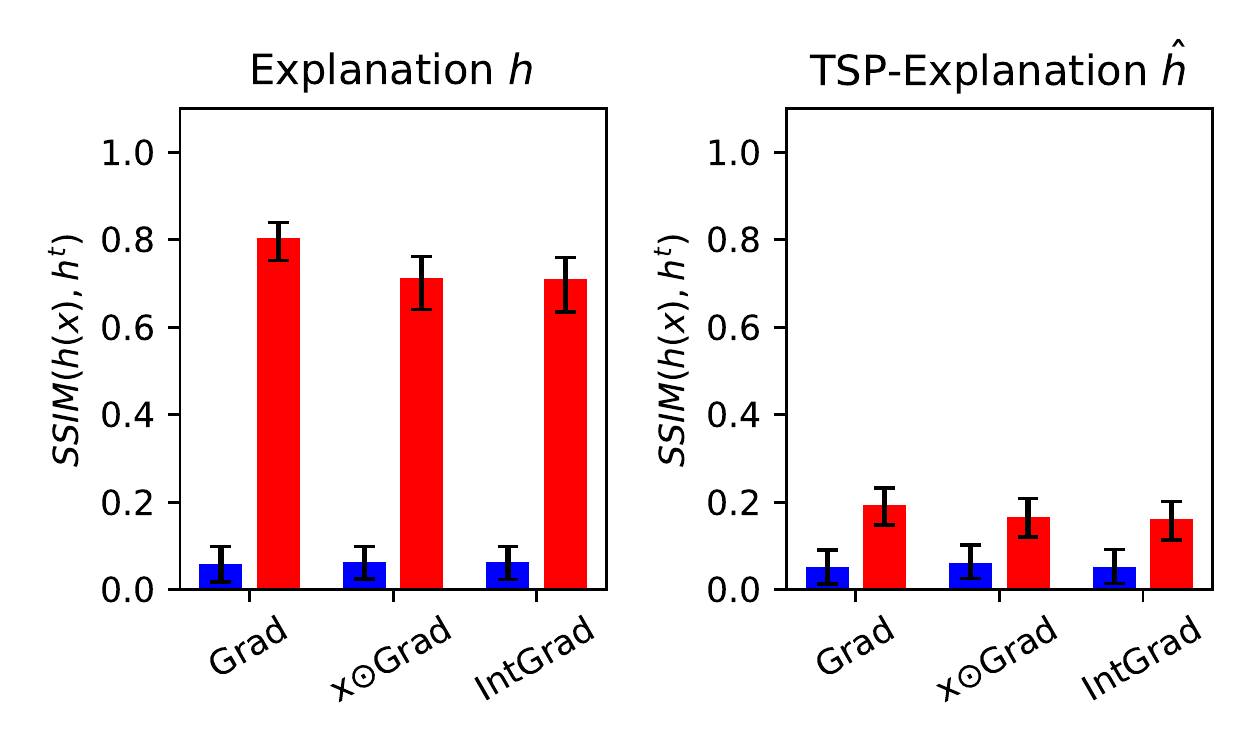}
  \caption{%
  \textbf{Left:} SSIM of the target map $h^t$ and explanations of \original{original model } and \manipulated{manipulated} respectively. %
  \textbf{Right:} Same as on the left but for \emph{tsp-explanations}. %
  The model $\tilde{g}$ was trained to manipulate the tsp-explanation, but this time with a higher weighting factor $\gamma = 9$. %
  Even with this more aggressive manipulation compared to the original experiment in Figure \ref{fig:quant}, tsp-explanations are considerably more robust than their unprojected counterparts on the left. %
  Colored bars show the median. %
  Errors denote the 25th and 75th percentile. %
  }
  \label{fig:quantHigh}
\end{figure}

All manipulated models on all datasets were trained using Adam \citeapp{kingma2015adam} by fine-tuning the original model with a fixed learning rate of $10^{-5}$ until convergence. 
We set the weighting factor $\gamma$ of the loss function \eqref{eq:loss} to $4$.
We use the same hyperparameters for manipulating tsp-explanations to ensure fair comparison.
To ensure our results do not depend on a specific weighting factor $\gamma$, we demonstrate the same experiment shown in Figure \ref{fig:quant} with $\gamma = 9$ in Figure \ref{fig:quantHigh}.

\begin{figure}[ht!]
  \centering
     \includegraphics[width=0.4\linewidth]{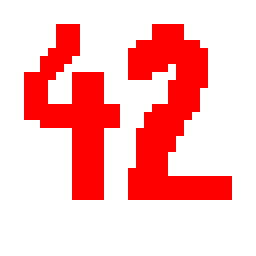}
   \caption{Image used as the target explanation to train the manipulated models.}
   \label{fig:target}
\end{figure}

\paragraph{Target Explanation:}
The target explanation map used in our experiments is shown in Figure~\ref{fig:target}.

\begin{figure}[ht!]
  \centering
     \includegraphics[width=0.49\linewidth]{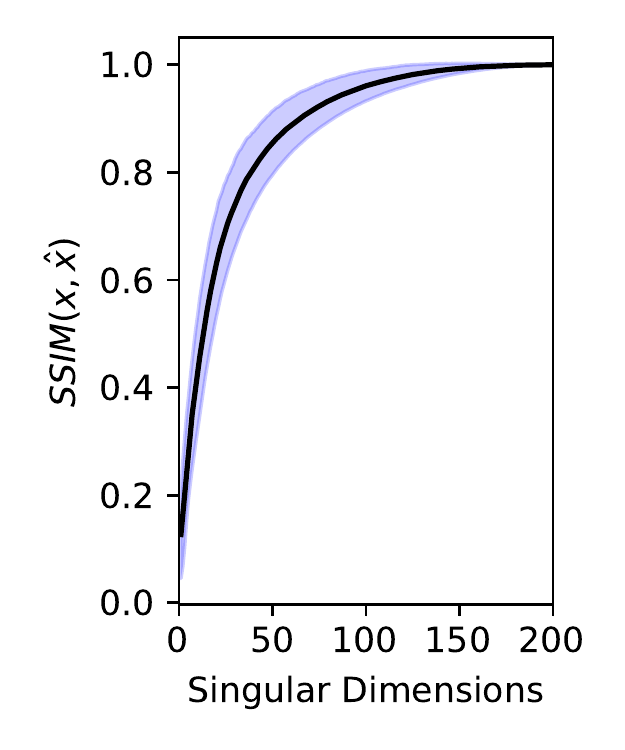}
     \includegraphics[width=0.49\linewidth]{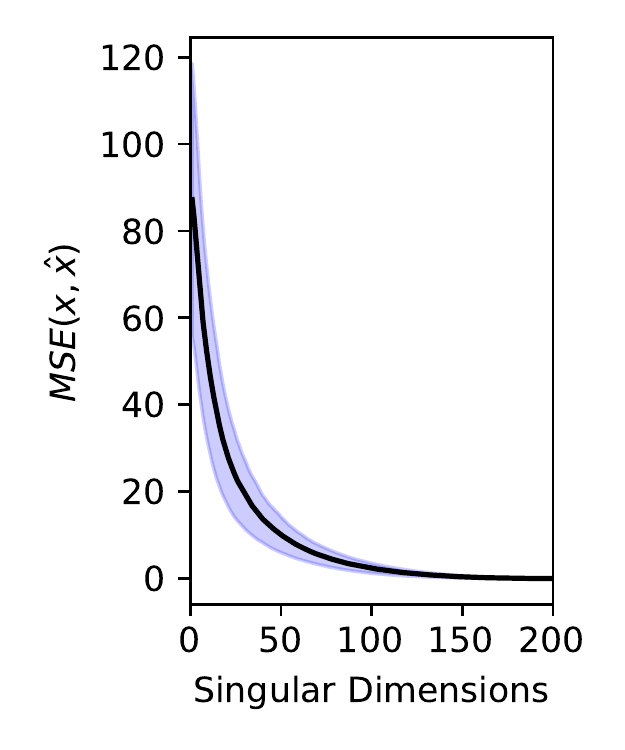}
   \caption{SSIM (left) and MSE (right) of original vs. reconstructed image from the tangent-space directions, ordered by number of used directions. Images are drawn from the full FashionMNIST test set. A total of 200 neighbours was used for each image. The black curve describes the median, while the surrounding blue area marks the space between the 25th and 75th percentiles.}
   \label{fig:reconst}
\end{figure}

\paragraph{Model Statistics:}
The accuracies, MSE and KL-divergence of the original and adversarially trained models are documented in Tables \ref{tab:acc}, \ref{tab:mse} and \ref{tab:kldiv} respectively.

\paragraph{Estimating Tangent Space:} In the following, we briefly summarize the procedure used to estimate tangent space for the various datasets.

\underline{MNIST, FMNIST:} We use the hyperplane method described in the main text. For a given data point, the nearest neighbours are taken only from the training set. The dimensionality of the hyperplane is chosen to be $30$. This number was tuned by ensuring that the data points are well reconstructed with respect to the MSE (which corresponds to the Euclidean distances, i.e. the natural metric on the embedding space $\mathbb{R}^D$), see Figure~\ref{fig:reconst}. The hyperplane is fitted using the nearest neighbours and the datapoint itself. Before fitting, all datapoints are normalized to have zero mean and a standard deviation of one.

\begin{figure}[ht!]
  \centering
     \includegraphics[width=0.5\linewidth]{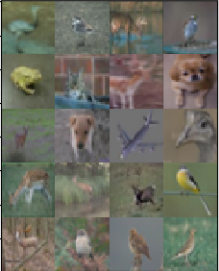}
   \caption{Nearest neighbours with respect to Euclidean distance for image in the top left-hand corner. Clearly, the neighbours are not local deformation of the image itself. As a result, the hyperplane method cannot be used for the CIFAR10 dataset.}
   \label{fig:nbrs}
\end{figure}

\underline{CIFAR10:} We use the autoencoder method described in the main text. This is because the manifold is not densly sampled enough for the hyperplane method, see Figure~\ref{fig:nbrs}. We normalize the data as described above and split it by class. A separate autoencoder is trained for each class for three epochs using the Adam optimizer with a learning rate of $0.001$. We use a same VQ-VAE architecture as in this  example\footnote{\url{https://github.com/deepmind/sonnet/blob/master/sonnet/examples/vqvae_example.ipynb}}. After training, the Jacobian $\tfrac{\partial D}{\partial z_e}(x)$ is calculated by backpropagation for each data sample $x$. We note that this could be sped up by forward-mode differentiation. We then perform an SVD-decomposition of the result and tune the number of singular components ensuring good reconstruction.

\begin{figure}[ht!]
  \centering
     \includegraphics[width=0.9\linewidth]{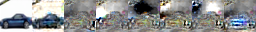}
   \caption{The input image is shown on the very left. Second image is the reconstruction from the tangent-space directions of which six are shown on the right.}
   \label{fig:tangent_vectors}
\end{figure}

\begin{table}
    \begin{tabular}{r|c c c}
        Method           & MNIST   & FashionMNIST & CIFAR10 \\ \hline
        Original         & $98.97$ & $94.72$      & $92.47$ \\
        Gradient         & $98.84$ & $94.58$      & $91.77$ \\
        x $\odot$ Grad   & $98.96$ & $94.48$      & $91.53$ \\
        IntGrad          & $98.95$ & $94.65$      & $91.62$ \\
        LRP              & $98.95$ & $94.68$      & $92.08$ \\
    \end{tabular}
    \caption{Accuracies of all models in percent.}
    \label{tab:acc}
\end{table}

\newcommand\temf{}

\begin{table}
    \begin{tabular}{r|c c c}
        Method         & MNIST          & FashionMNIST           & CIFAR10                \\ \hline
        Gradient       & $120.54 \temf$ & $11.13 \temf$ & $838.80 \temf$ \\
        x $\odot$ Grad & $114.29 \temf$ & $15.07 \temf$ & $933.38 \temf$ \\
        IntGrad        & $128.03 \temf$ & $13.04 \temf$ & $707.52 \temf$ \\
        LRP            & $119.08 \temf$ & $ 3.76 \temf$ & $647.45 \temf$ \\
    \end{tabular}
    \caption{MSE$\times 10^{5}$ of model outputs $g(x)$ and $\tilde{g}(x)$ after final softmax.}
    \label{tab:mse}
\end{table}

\begin{table}
    \begin{tabular}{r|c c c}
        Method         & MNIST         & FashionMNIST & CIFAR10      \\ \hline
        Gradient       & $ 1.21 \temf$ & $1.99 \temf$ & $8.39 \temf$ \\
        x $\odot$ Grad & $ 1.14 \temf$ & $2.06 \temf$ & $9.34 \temf$ \\
        IntGrad        & $ 1.50 \temf$ & $2.00 \temf$ & $6.30 \temf$ \\
        LRP            & $ 1.19 \temf$ & $1.19 \temf$ & $6.88 \temf$ \\
    \end{tabular}
    \caption{Mean KL-Divergence$\times 10^{3}$ between models $g$ and $\tilde{g}$.}
    \label{tab:kldiv}
\end{table}

\subsection{FashionMNIST}
\subsubsection{Additional Heatmaps}
\FloatBarrier

\begin{figure}[ht!]
  \centering
  \includegraphics[width=1\linewidth]{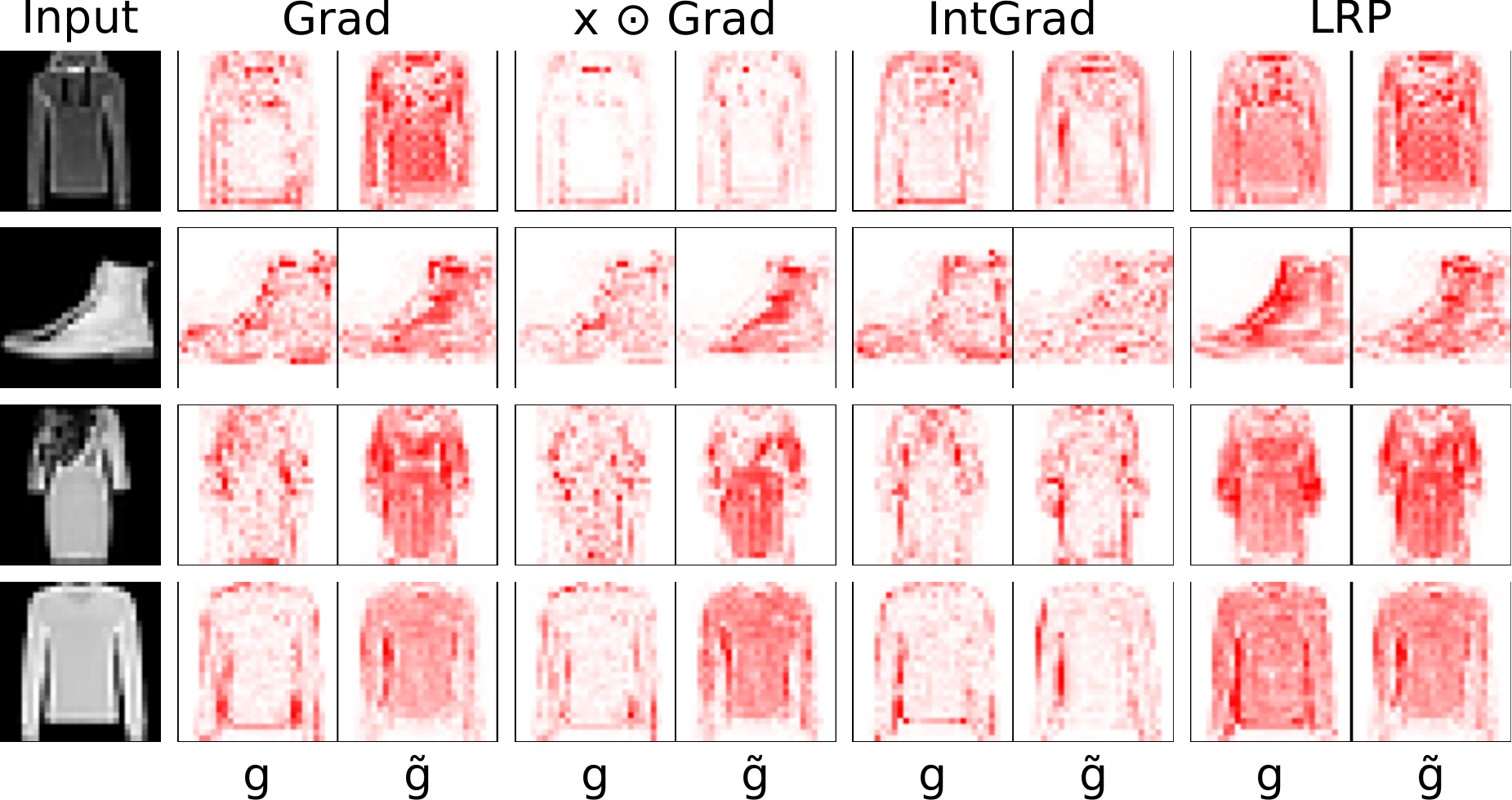}
  \caption{%
    Projected explanations from the original model $g$ (left) and the manipulated model $\tilde{g}$ (right) where the projected heatmaps were attacked for various images from the FashionMNIST test set.%
  }
  \label{fig:defended}
\end{figure}

\FloatBarrier
\subsubsection{Additional Distance Metrics for Quantitative Comparison}

\begin{figure}[ht!]
  \centering
  \includegraphics[width=1.\linewidth]{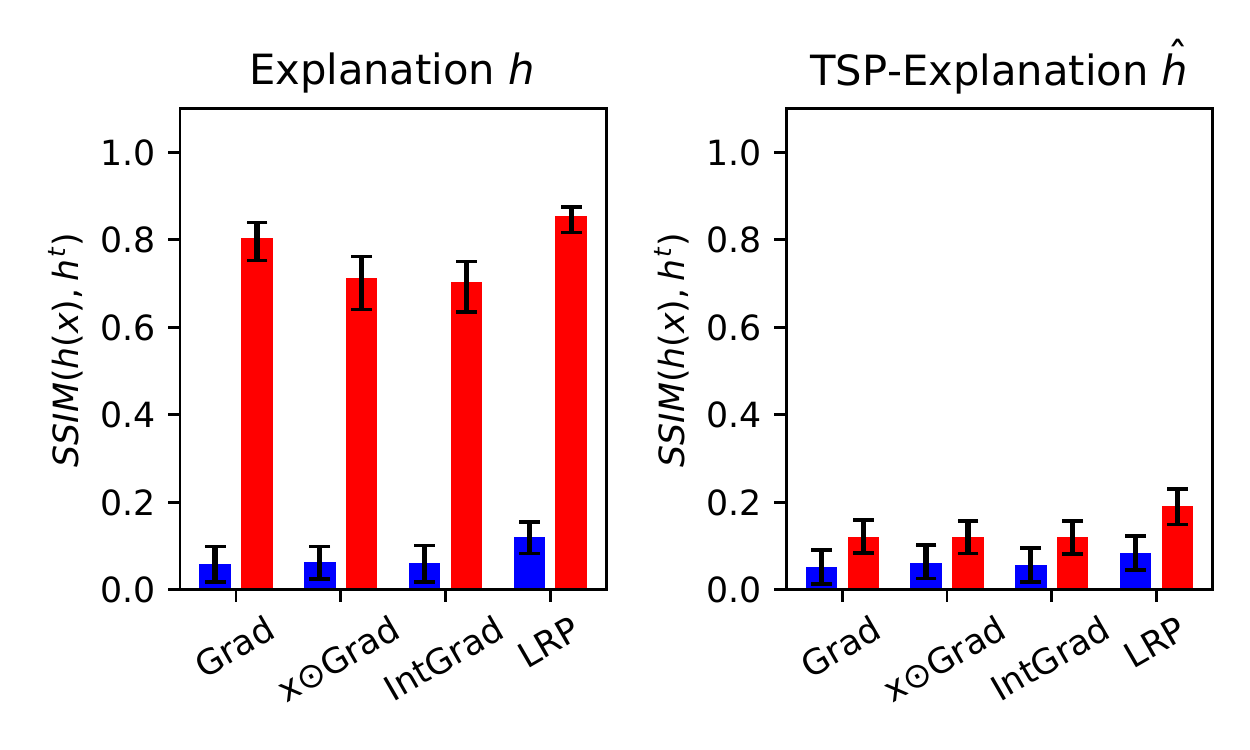}
  \caption{%
      Median of SSIM of left: $h_g(x)$ (blue) and $h_{\tilde{g}}$ (red), %
      right: $\hat{h}_g(x)$ (blue) and $\hat{h}_{\tilde{g}}$ (red) where $h(x)$ was manipulated,  on FashionMNIST. %
  }

  \label{fig:quantProj}
\end{figure}

\begin{figure}[ht!]
  \centering
  \includegraphics[width=1.\linewidth]{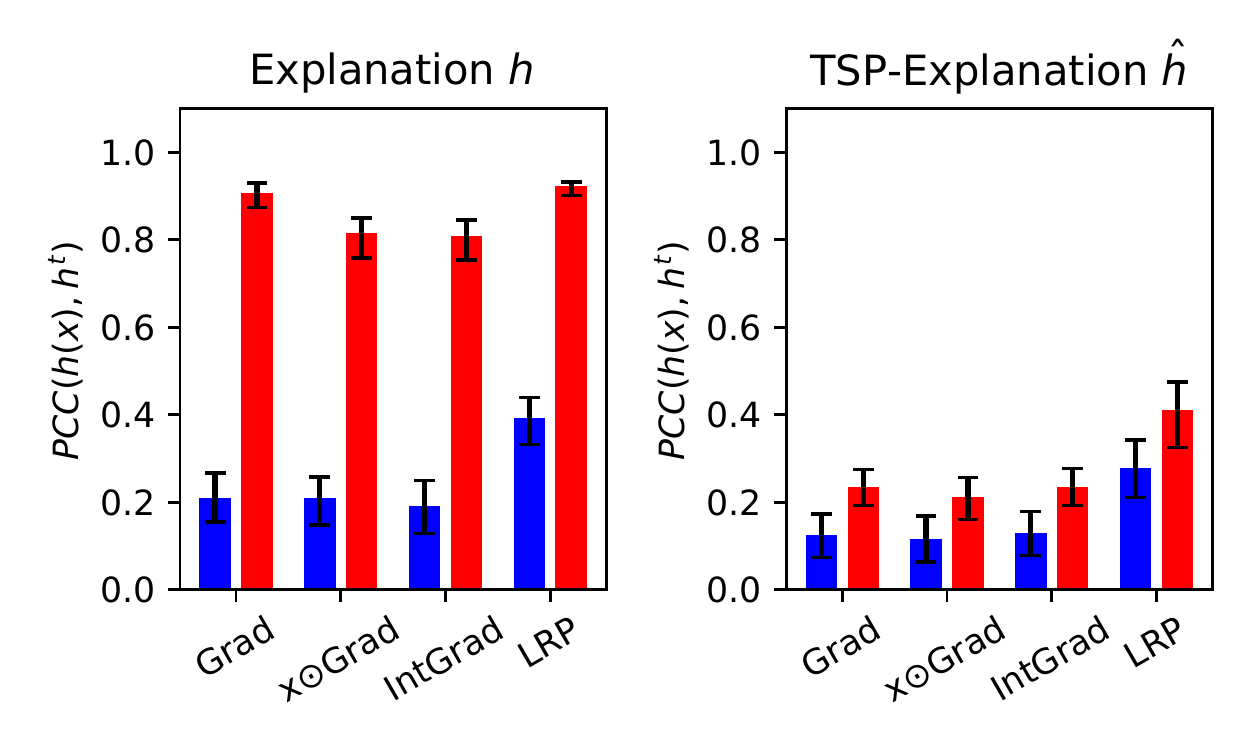}
  \caption{%
      Median of PCC of $\hat{h}_g(x)$ (blue) and $\hat{h}_{\tilde{g}}$ (red) on FashionMNIST where $h(x)$ was manipulated. %
  }
\end{figure}

\begin{figure}[ht!]
  \centering
  \includegraphics[width=1.\linewidth]{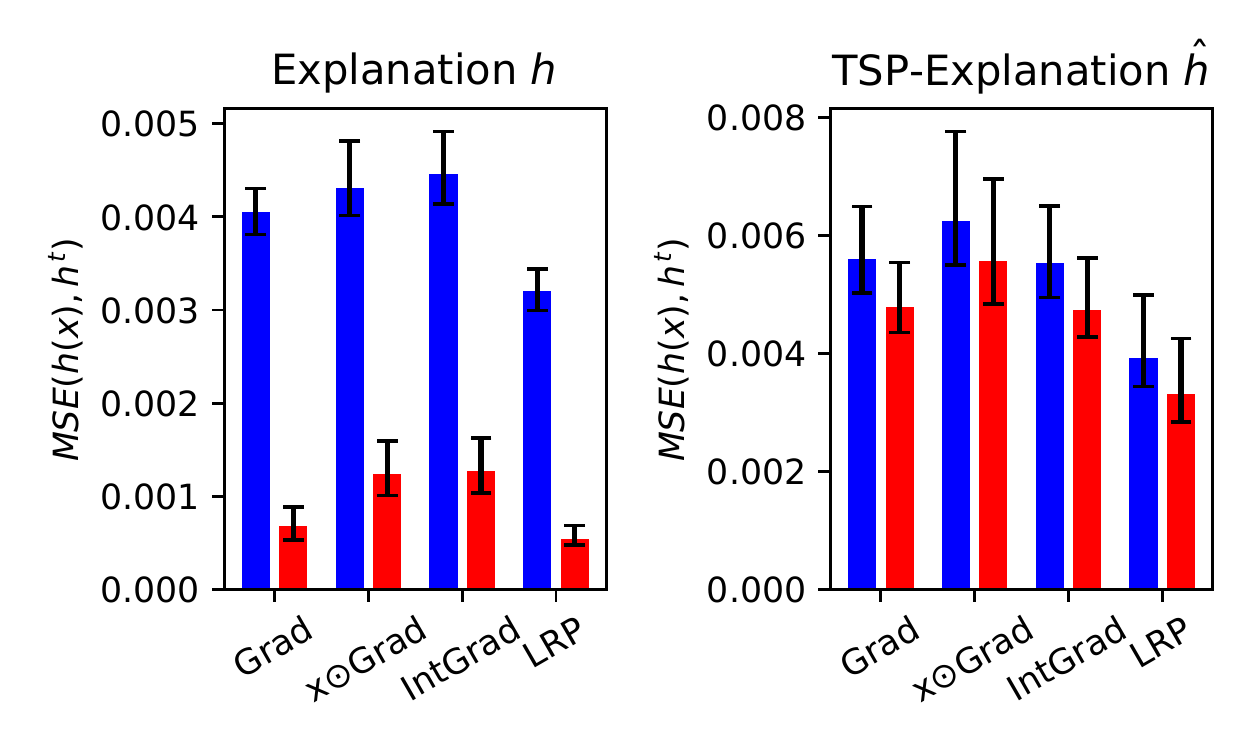}
  \caption{%
      Median of MSE of $\hat{h}_g(x)$ (blue) and $\hat{h}_{\tilde{g}}$ (red) on FashionMNIST where $h(x)$ was manipulated. %
  }
\end{figure}

\begin{figure}[ht!]
  \centering
  \includegraphics[width=1.\linewidth]{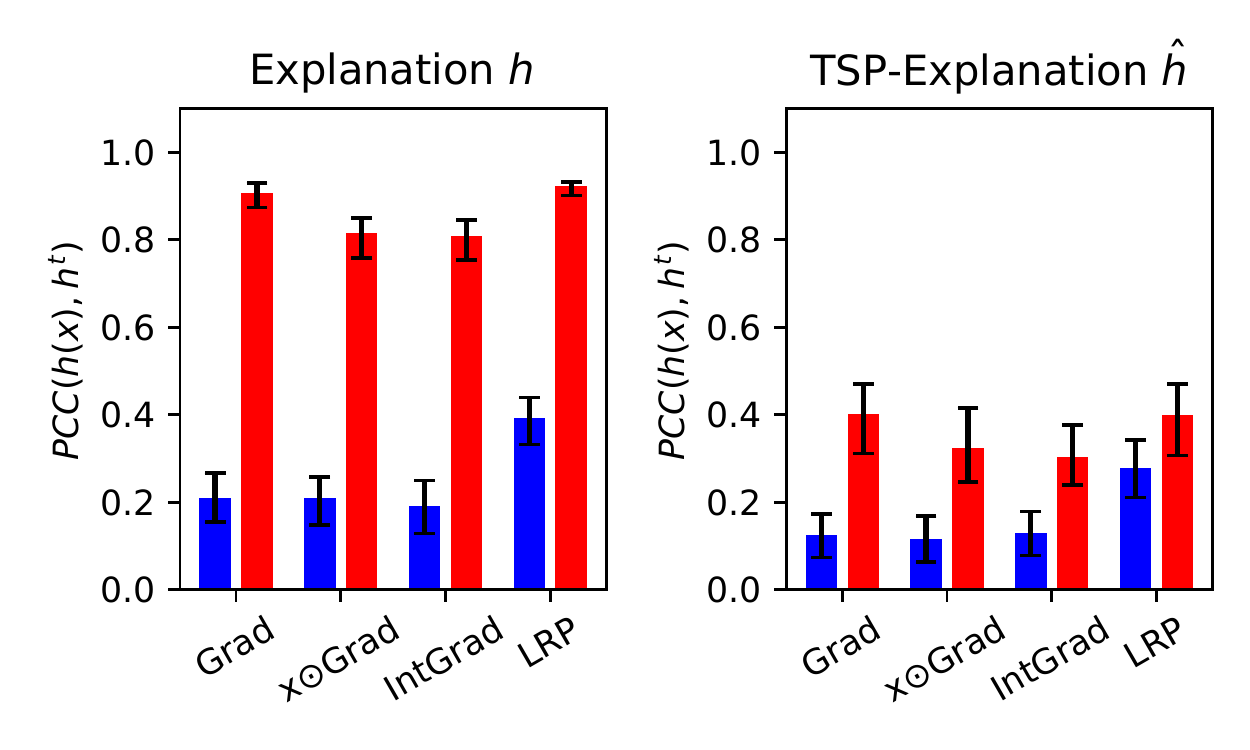}
  \caption{%
    Median of PCC of $\hat{h}_g(x)$ (blue) and $\hat{h}_{\tilde{g}}$ (red) on FashionMNIST where $\hat{h}(x)$ was manipulated. %
  }
\end{figure}

\begin{figure}[ht!]
  \centering
  \includegraphics[width=1.\linewidth]{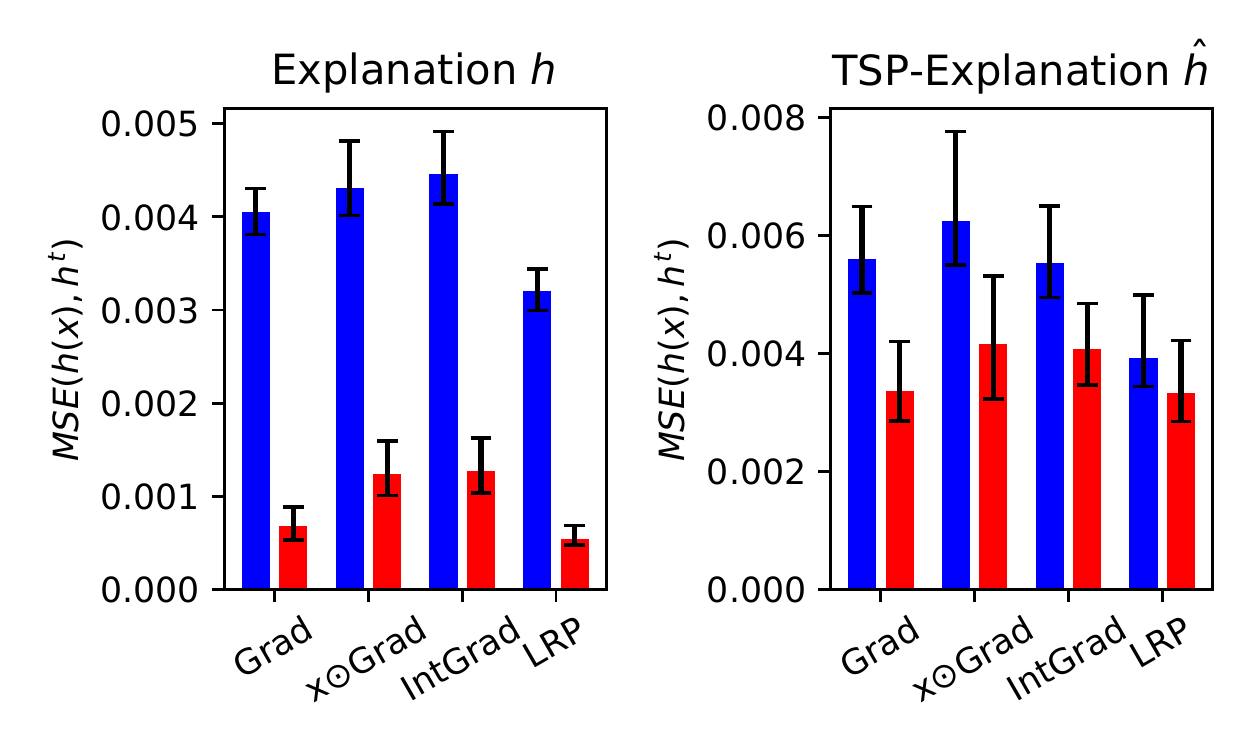}
  \caption{%
    Median of MSE of $\hat{h}_g(x)$ (blue) and $\hat{h}_{\tilde{g}}$ (red) on FashionMNIST where $\hat{h}(x)$ was manipulated. %
  }
\end{figure}

\FloatBarrier
\subsection{MNIST}
\subsubsection{Heatmaps}
\begin{figure}[ht!]
    \centering
    \includegraphics[width=1\linewidth]{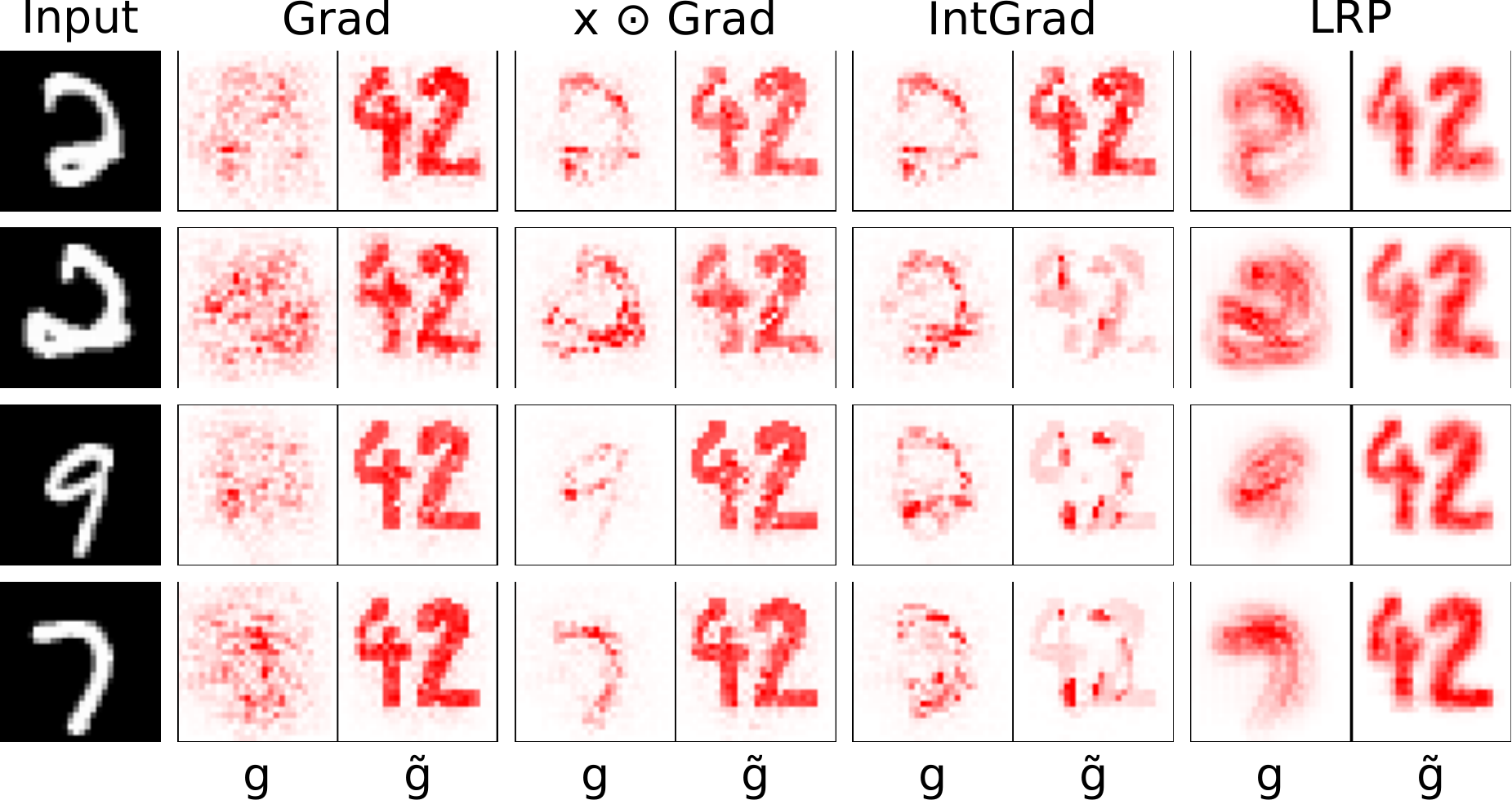}
    \caption{%
        Example explanations from the original model $g$ (left) and the manipulated model $\tilde{g}$ (right) for various images from the MNIST test set.}
\end{figure}

\begin{figure}[ht!]
  \centering
  \includegraphics[width=1\linewidth]{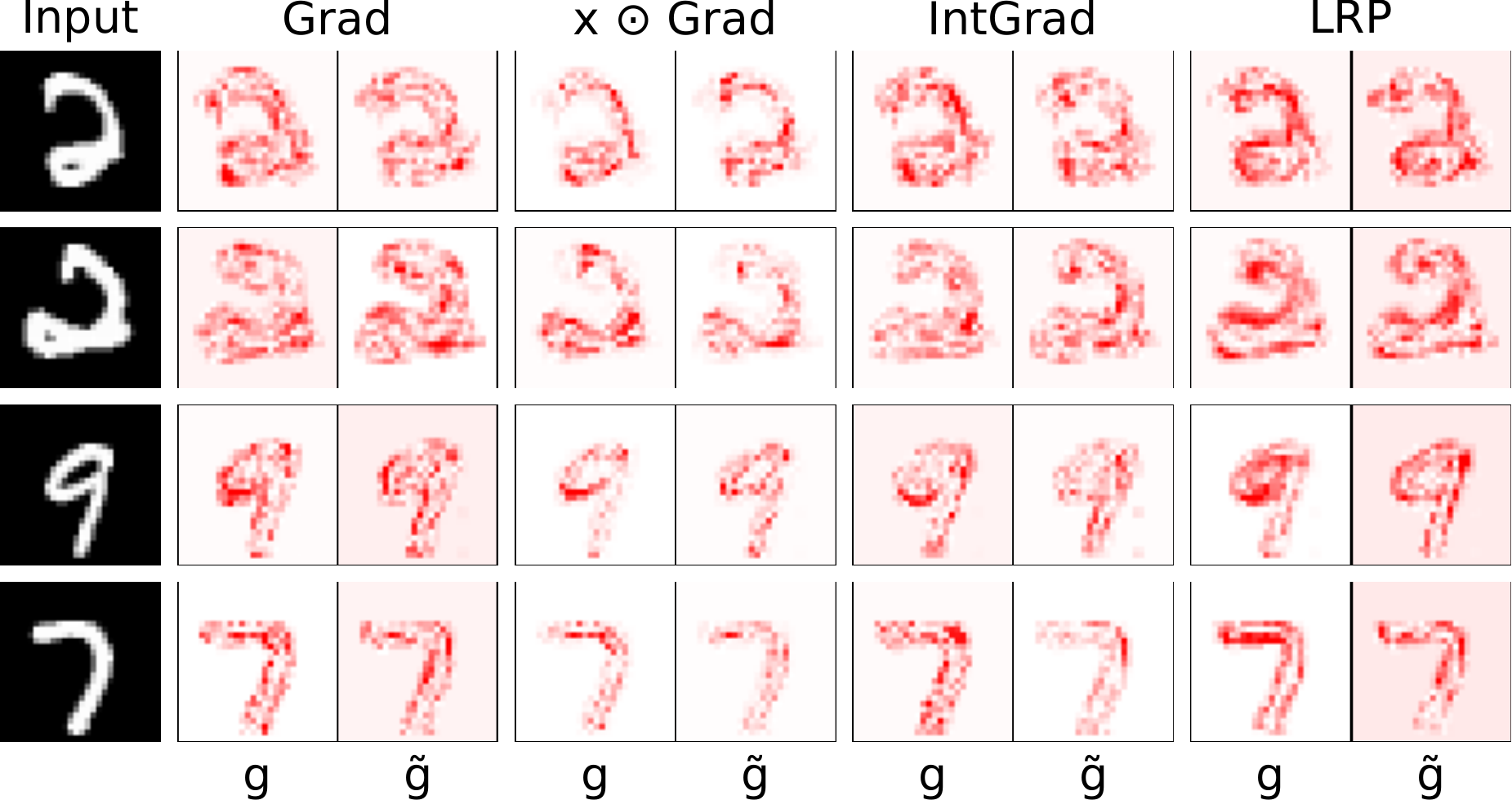}
  \caption{%
      Example tsp-explanations from the original model $g$ (left) and the manipulated model $\tilde{g}$ (right) for various images from the MNIST test set.%
  }
\end{figure}

\begin{figure}[ht!]
  \centering
  \includegraphics[width=1\linewidth]{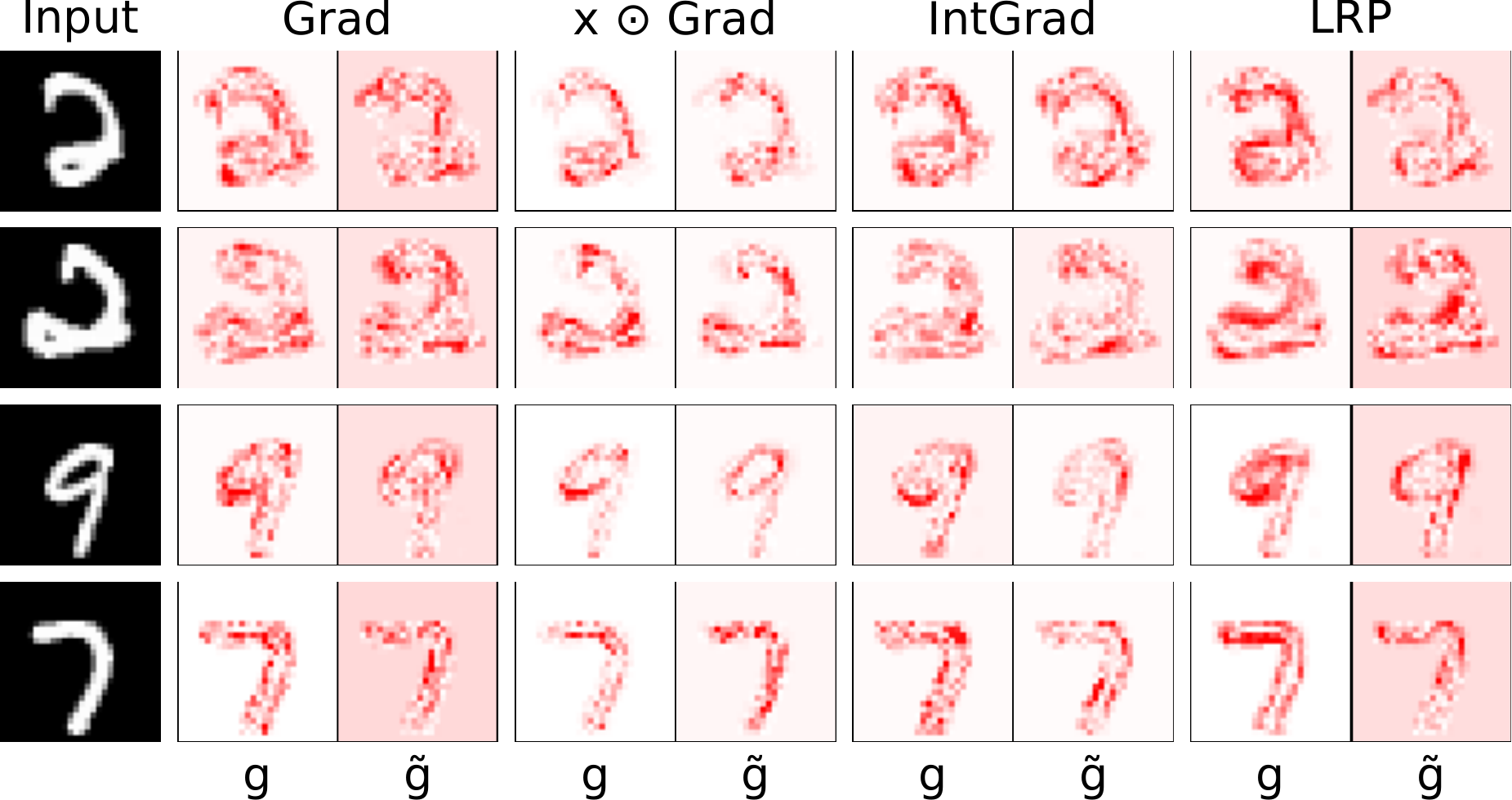}
  \caption{%
      Example tsp-explanations from the original model $g$ (left) and the manipulated model $\tilde{g}$ (right) where the projected heatmaps were attacked for various images from the MNIST test set.%
  }
\end{figure}

\FloatBarrier
\subsubsection{Quantitative Comparison}

\begin{figure}[ht!]
  \centering
  \includegraphics[width=1.\linewidth]{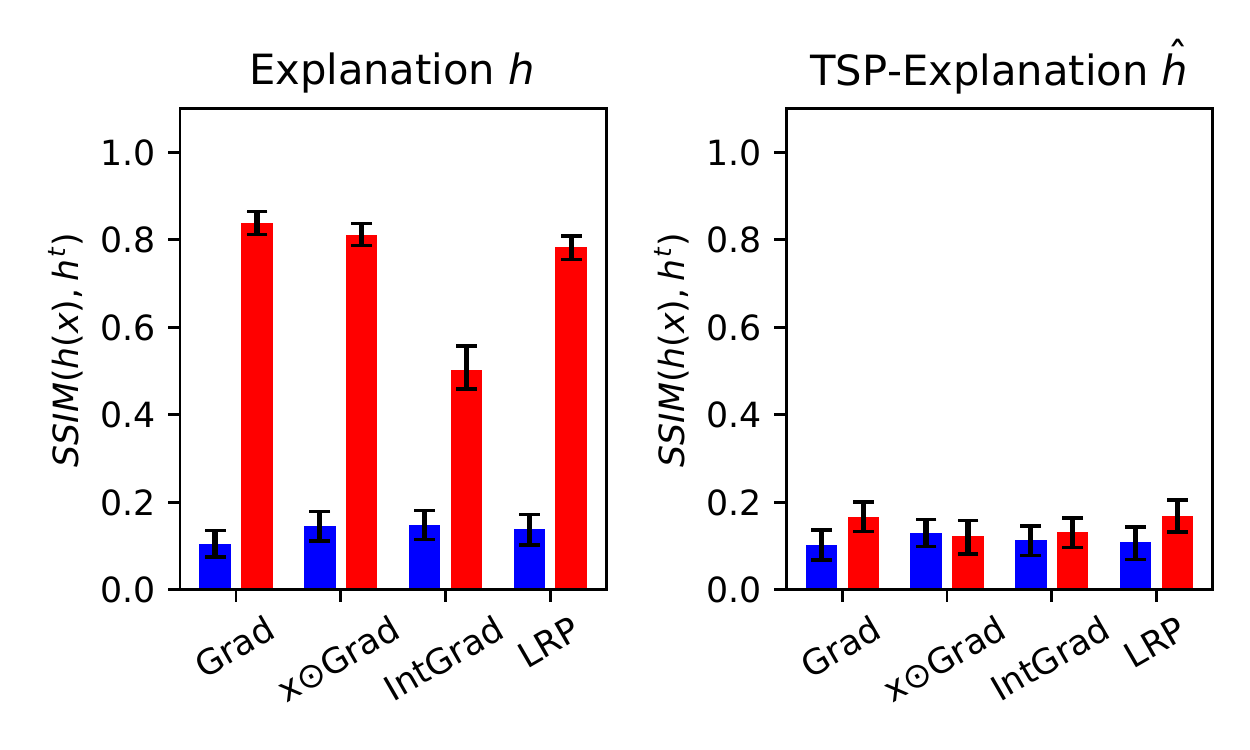}
  \caption{%
      Median of SSIM of left: $h_g(x)$ (blue) and $h_{\tilde{g}}$ (red), %
      right: $\hat{h}_g(x)$ (blue) and $\hat{h}_{\tilde{g}}$ (red) where $\hat{h}(x)$ was manipulated, on MNIST. %
  }

\end{figure}

\begin{figure}[ht!]
  \centering
  \includegraphics[width=1.\linewidth]{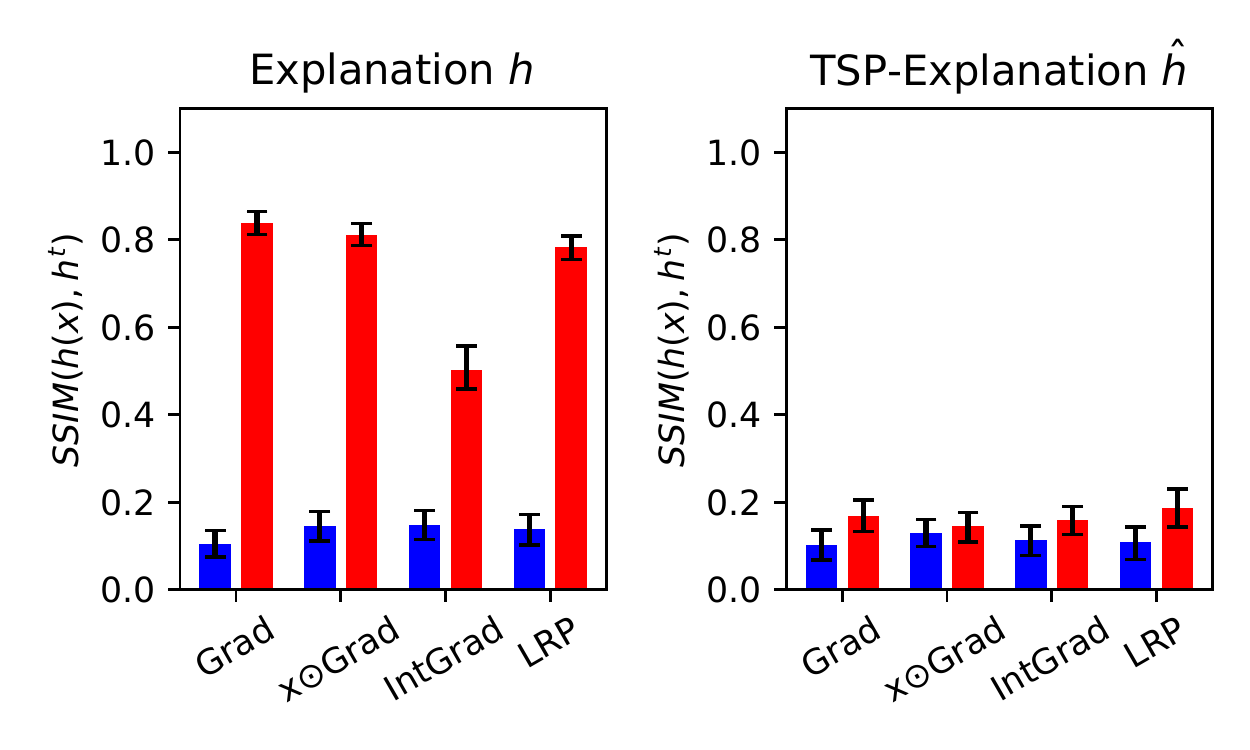}
  \caption{%
      Median of SSIM of left: $h_g(x)$ (blue) and $h_{\tilde{g}}$ (red), %
      right: $\hat{h}_g(x)$ (blue) and $\hat{h}_{\tilde{g}}$ (red) where $h(x)$ was manipulated, on MNIST. %
  }

\end{figure}

\begin{figure}[ht!]
  \centering
  \includegraphics[width=1.\linewidth]{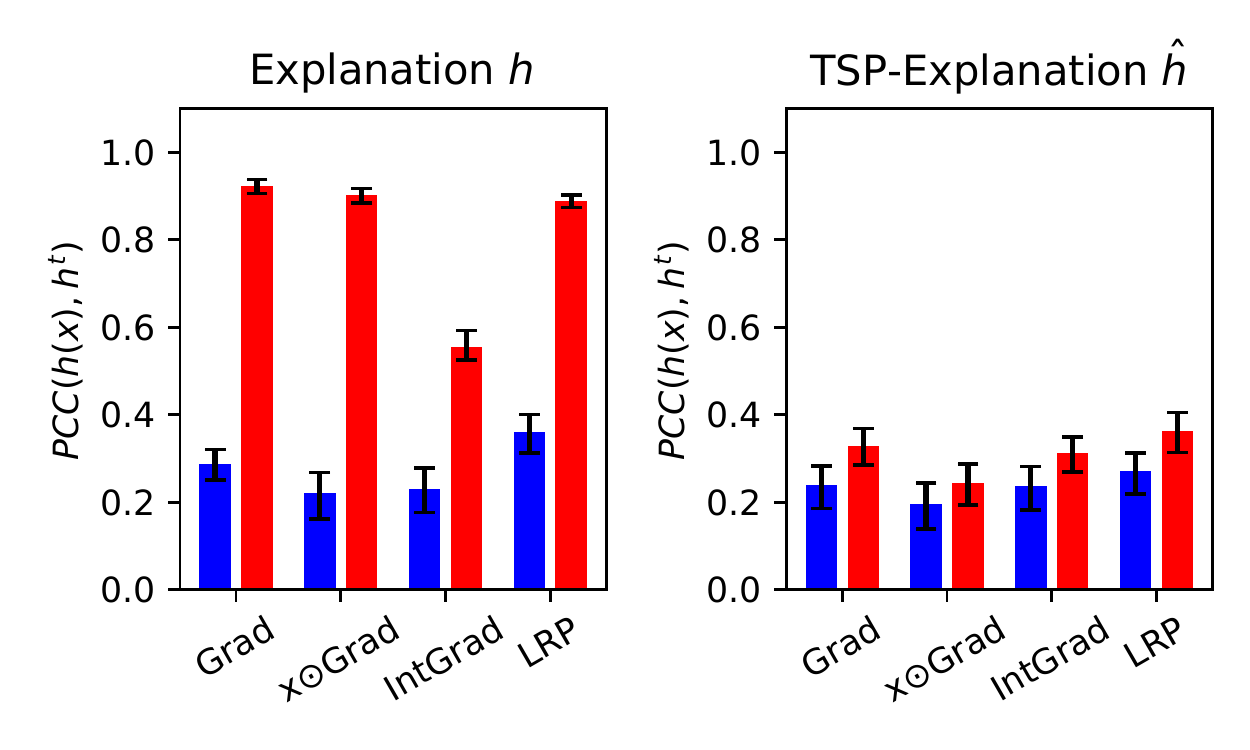}
  \caption{%
      Median of PCC of $\hat{h}_g(x)$ (blue) and $\hat{h}_{\tilde{g}}$ (red) on MNIST where $h(x)$ was manipulated. %
  }
\end{figure}

\begin{figure}
  \centering
  \includegraphics[width=1.\linewidth]{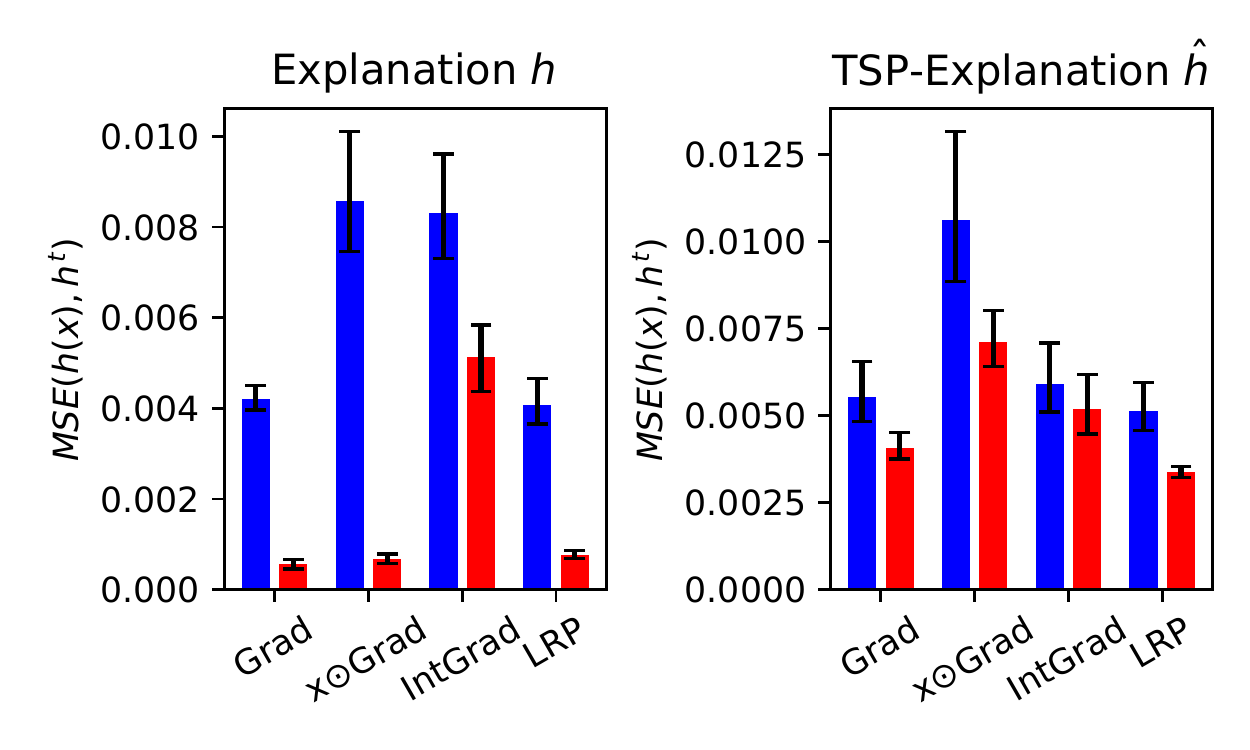}
  \caption{%
      Median of MSE of $\hat{h}_g(x)$ (blue) and $\hat{h}_{\tilde{g}}$ (red) on MNIST where $h(x)$ was manipulated. %
  }
\end{figure}

\begin{figure}[ht!]
  \centering
  \includegraphics[width=1.\linewidth]{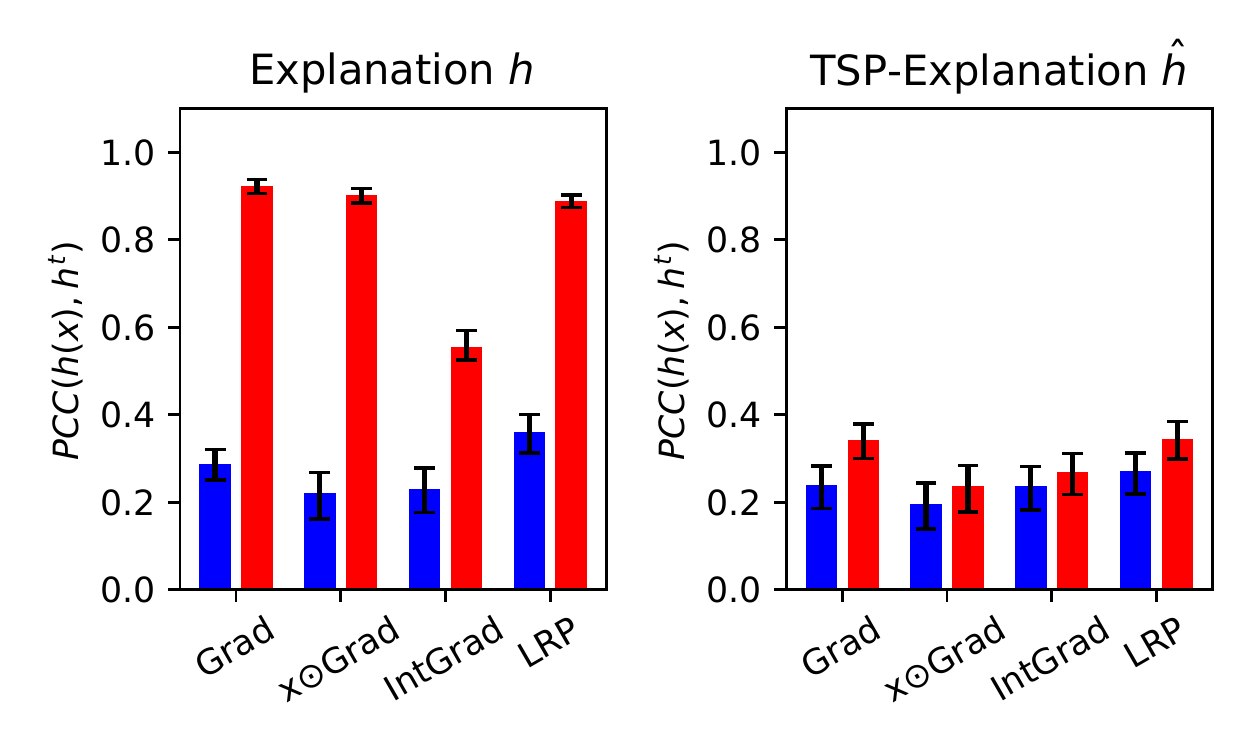}
  \caption{%
      Median of PCC of $\hat{h}_g(x)$ (blue) and $\hat{h}_{\tilde{g}}$ (red) on MNIST where $\hat{h}(x)$ was manipulated. %
  }
\end{figure}

\begin{figure}[ht!]
  \centering
  \includegraphics[width=1.\linewidth]{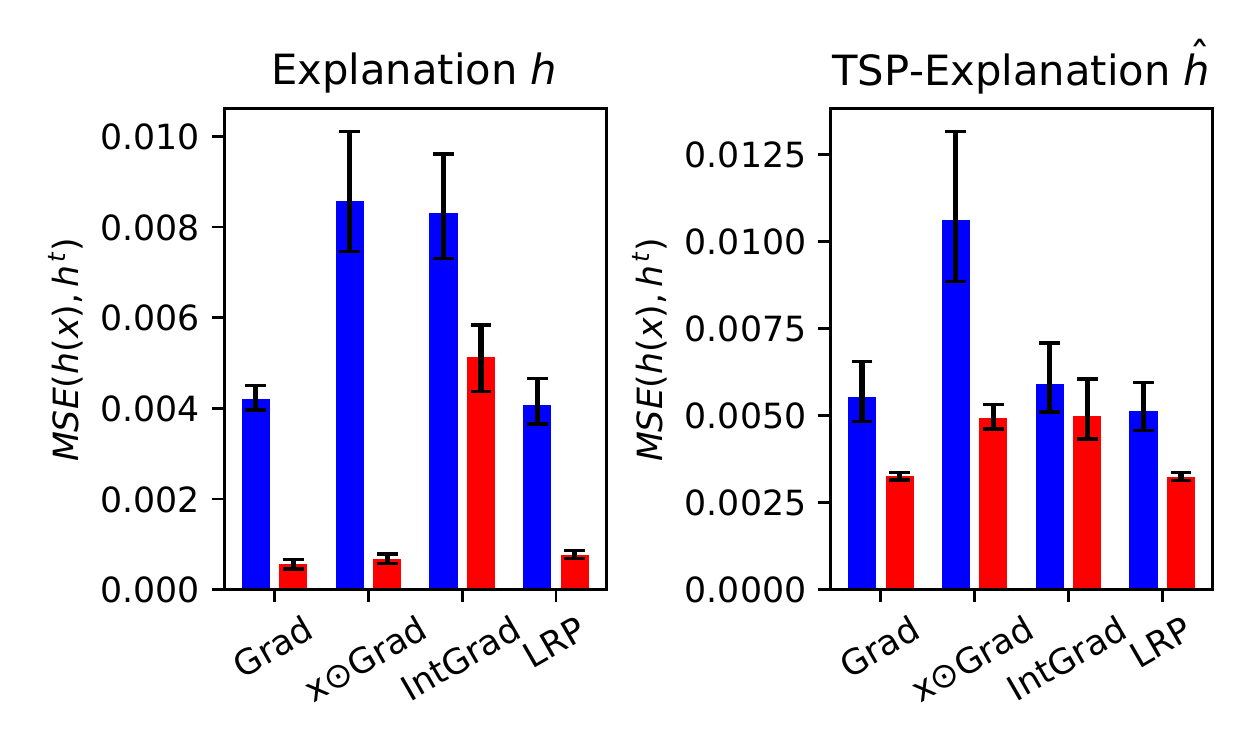}
  \caption{%
      Median of MSE of $\hat{h}_g(x)$ (blue) and $\hat{h}_{\tilde{g}}$ (red) on MNIST where $\hat{h}(x)$ was manipulated. %
  }
\end{figure}

\FloatBarrier
\subsection{CIFAR10}
\subsubsection{Heatmaps}
\begin{figure}[ht!]
    \centering
    \includegraphics[width=1\linewidth]{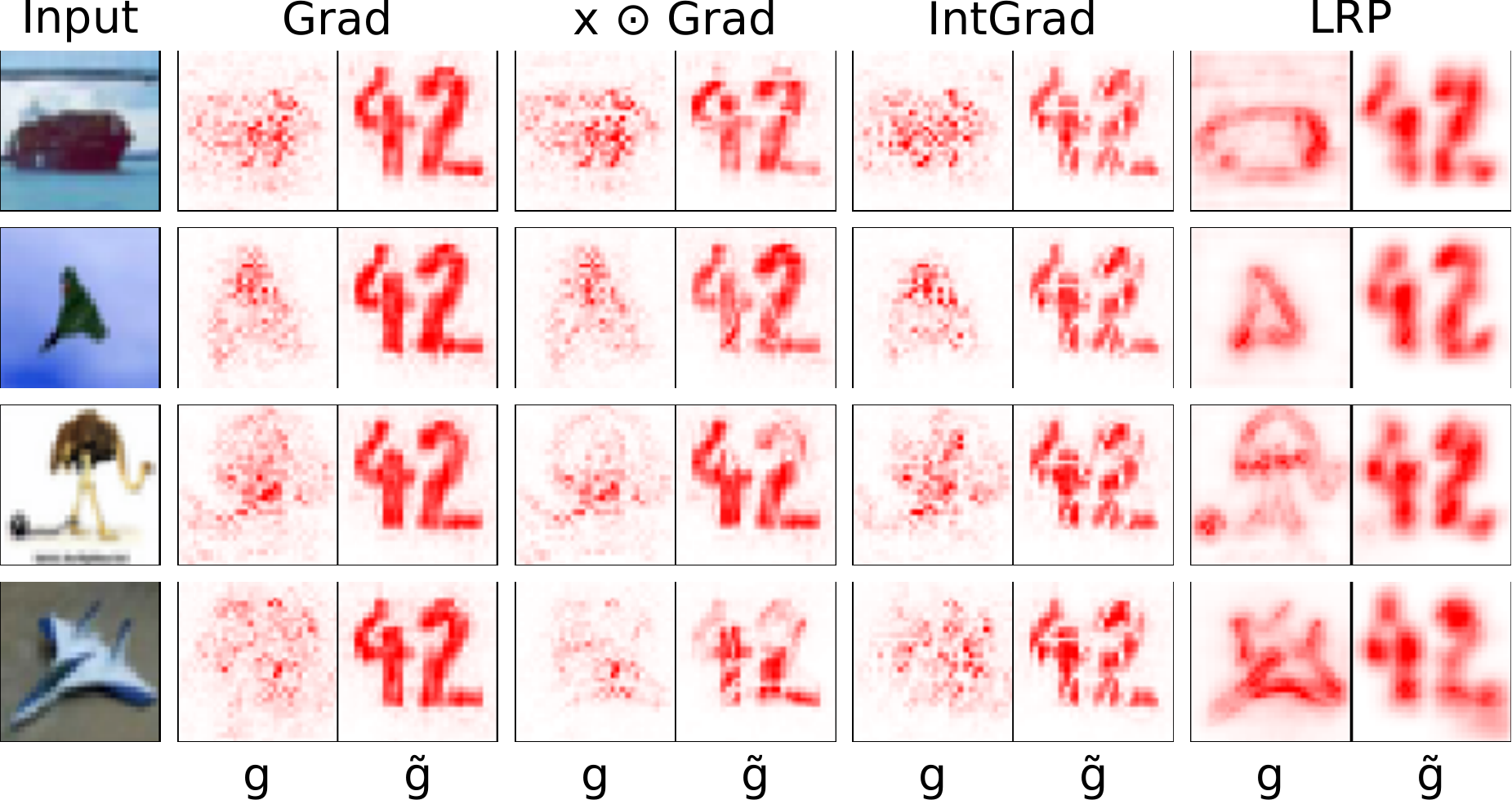}
    \caption{%
        Example explanations from the original model $g$ (left) and the manipulated model $\tilde{g}$ (right) for various images from the CIFAR10 test set.}
\end{figure}

\begin{figure}[ht!]
  \centering
  \includegraphics[width=1\linewidth]{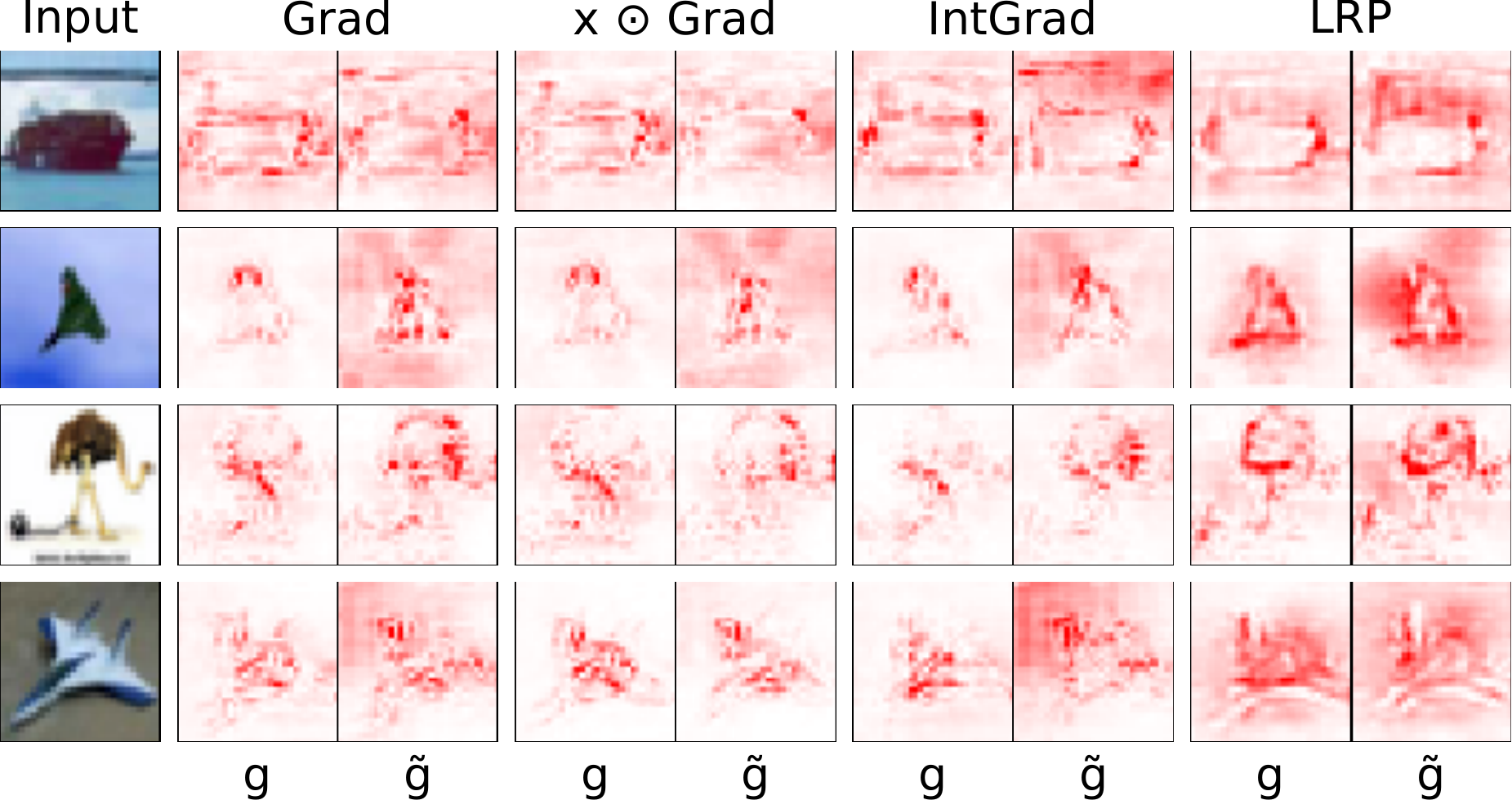}
  \caption{%
      Example tsp-explanations from the original model $g$ (left) and the manipulated model $\tilde{g}$ (right) for various images from the CIFAR10 test set.%
  }
\end{figure}

\begin{figure}[ht!]
  \centering
  \includegraphics[width=1\linewidth]{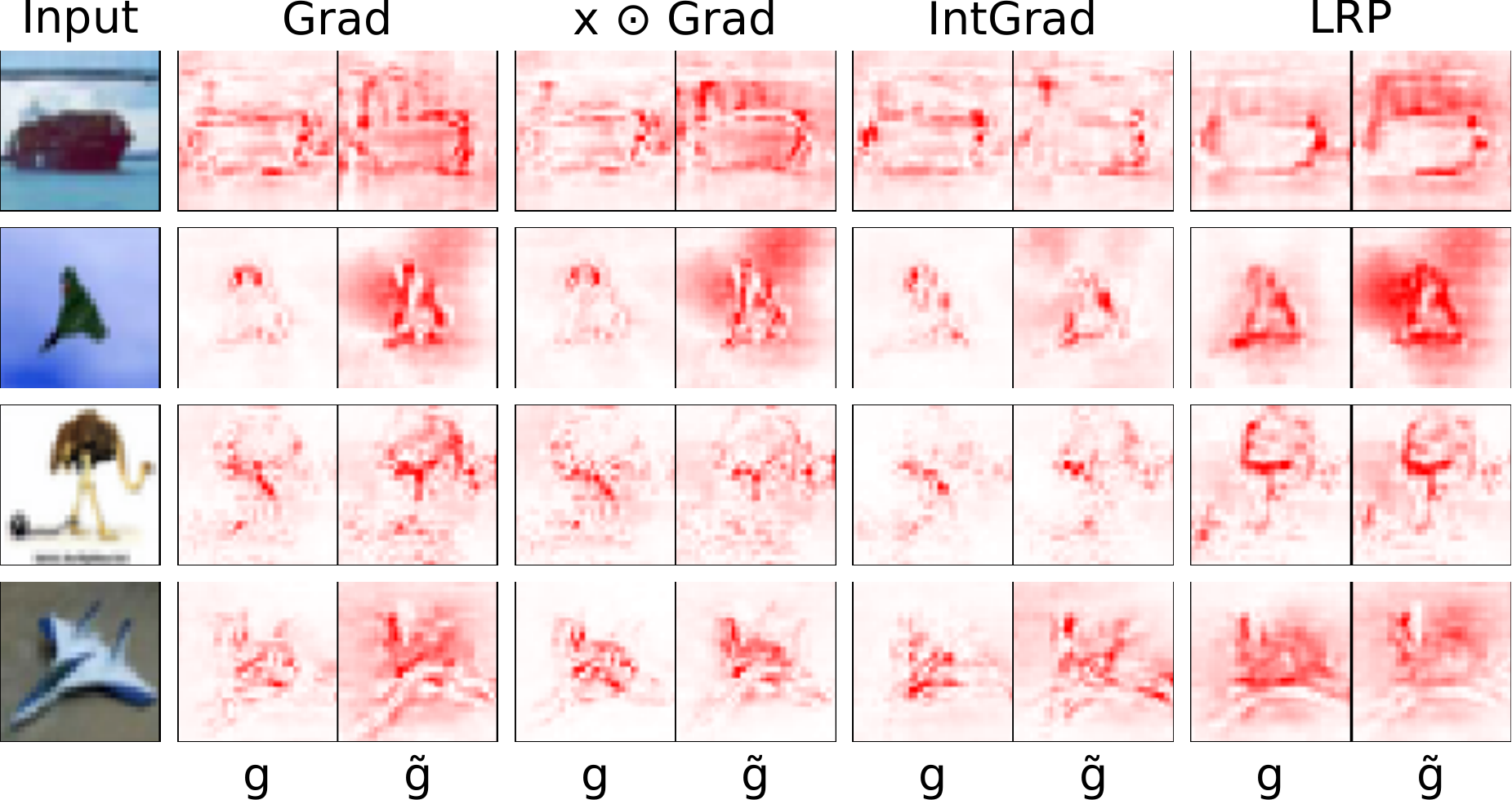}
  \caption{%
      Example tsp-explanations from the original model $g$ (left) and the manipulated model $\tilde{g}$ (right) where the projected heatmaps were attacked for various images from the CIFAR10 test set.%
  }
\end{figure}

\FloatBarrier
\subsubsection{Quantitative Comparison}

\begin{figure}[ht!]
  \centering
  \includegraphics[width=1.\linewidth]{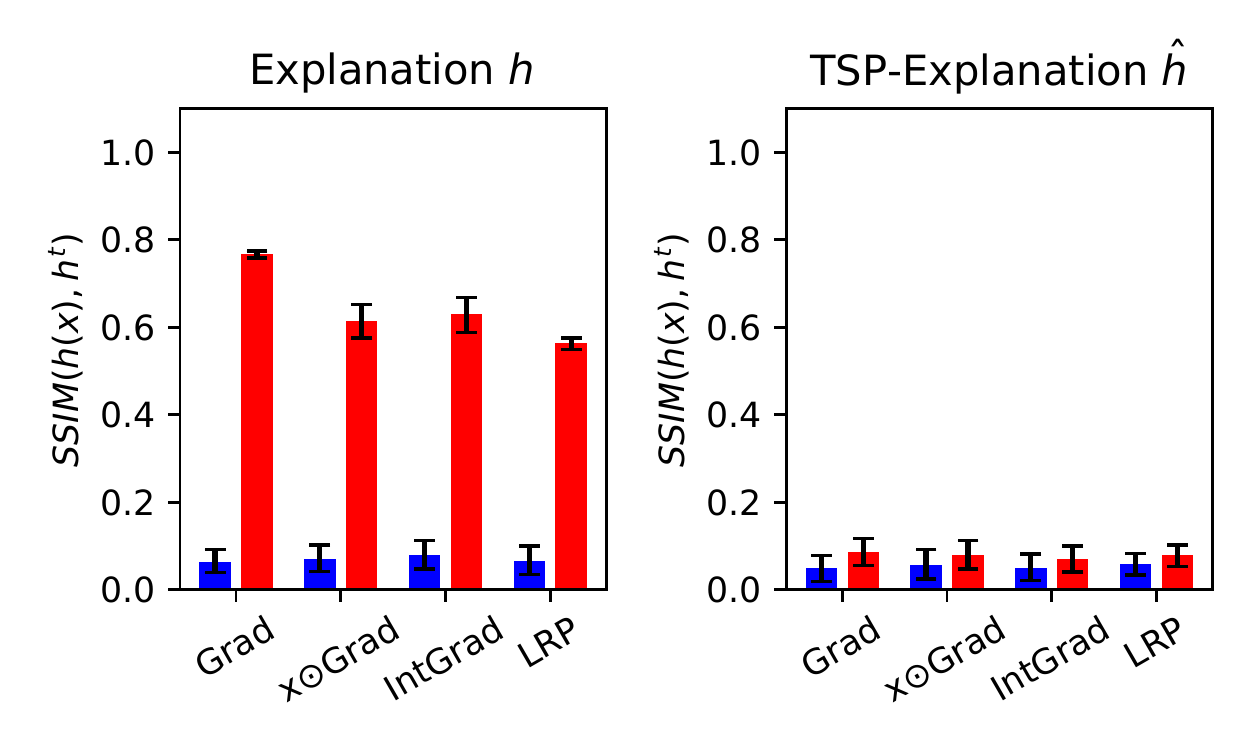}
  \caption{%
      Median of SSIM of left: $h_g(x)$ (blue) and $h_{\tilde{g}}$ (red), %
      right: $\hat{h}_g(x)$ (blue) and $\hat{h}_{\tilde{g}}$ (red) where $\hat{h}(x)$ was manipulated, on CIFAR10. %
  }

\end{figure}

\begin{figure}[ht!]
  \centering
  \includegraphics[width=1.\linewidth]{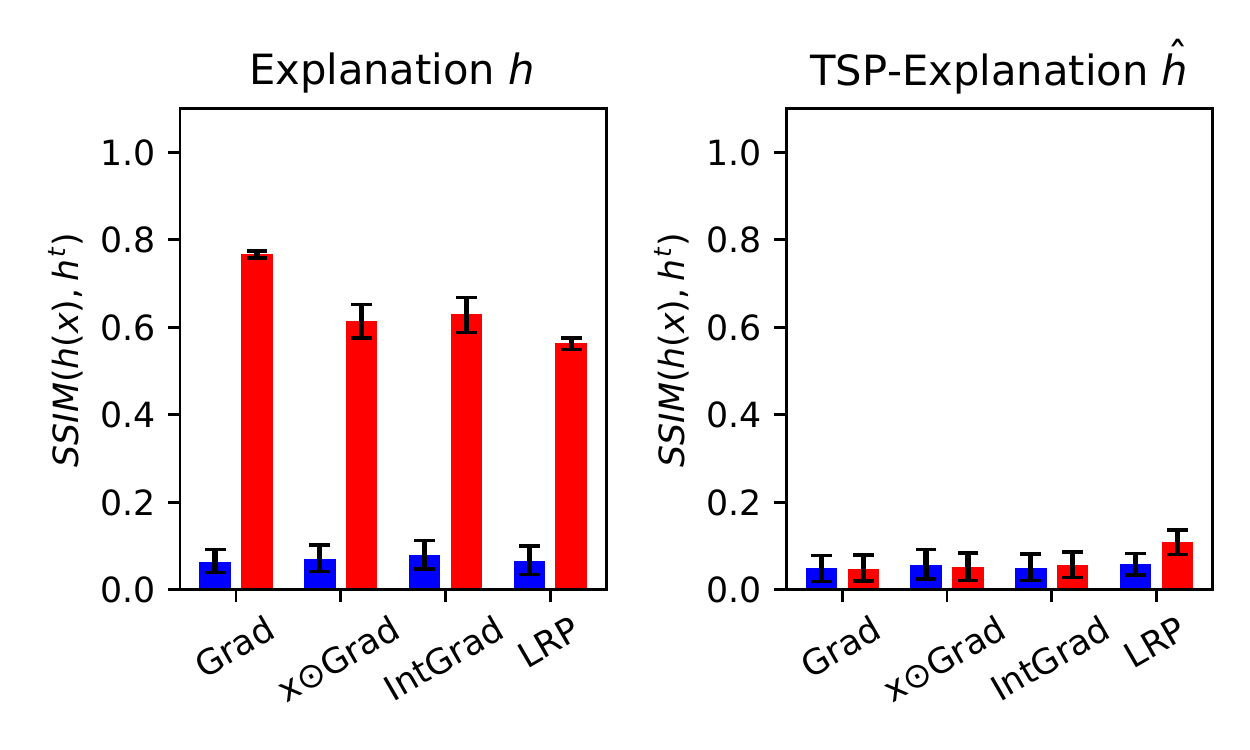}
  \caption{%
      Median of SSIM of left: $h_g(x)$ (blue) and $h_{\tilde{g}}$ (red), %
      right: $\hat{h}_g(x)$ (blue) and $\hat{h}_{\tilde{g}}$ (red) where $h(x)$ was manipulated, on CIFAR10. %
  }

\end{figure}

\begin{figure}[ht!]
  \centering
  \includegraphics[width=1.\linewidth]{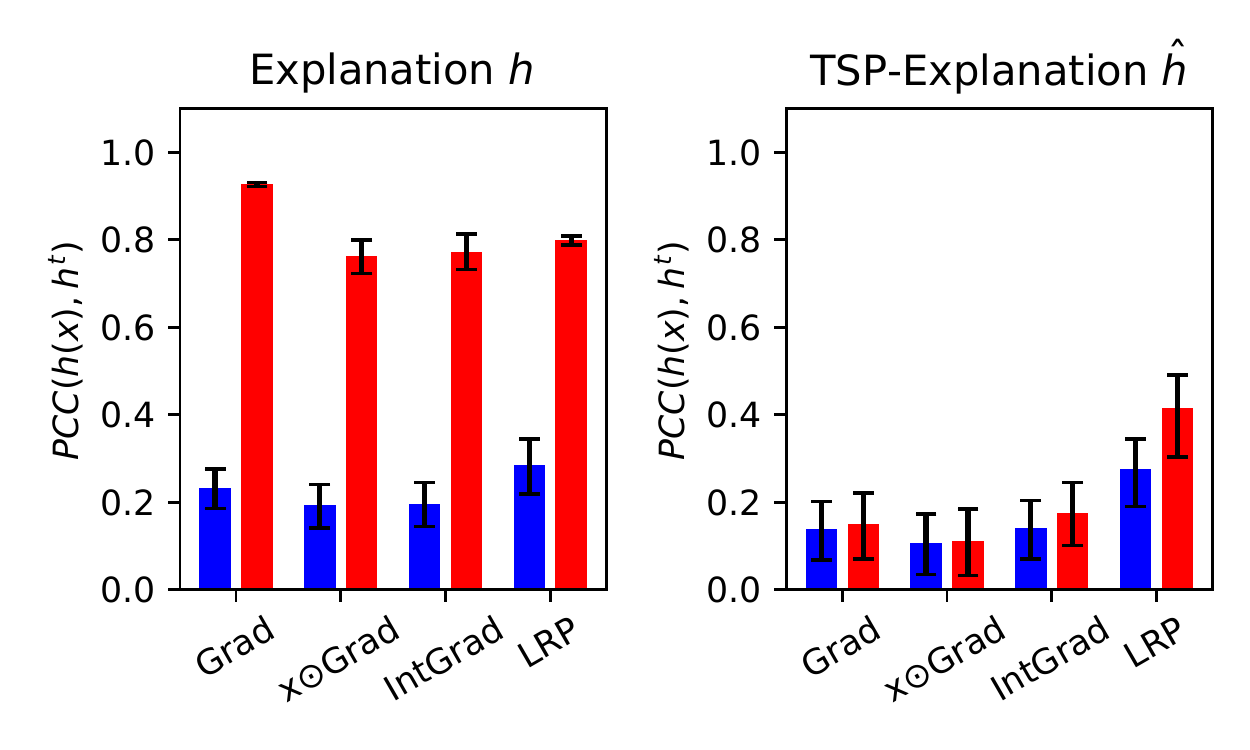}
  \caption{%
      Median of PCC of $\hat{h}_g(x)$ (blue) and $\hat{h}_{\tilde{g}}$ (red) on CIFAR10 where $h(x)$ was manipulated. %
  }
\end{figure}

\begin{figure}
  \centering
  \includegraphics[width=1.\linewidth]{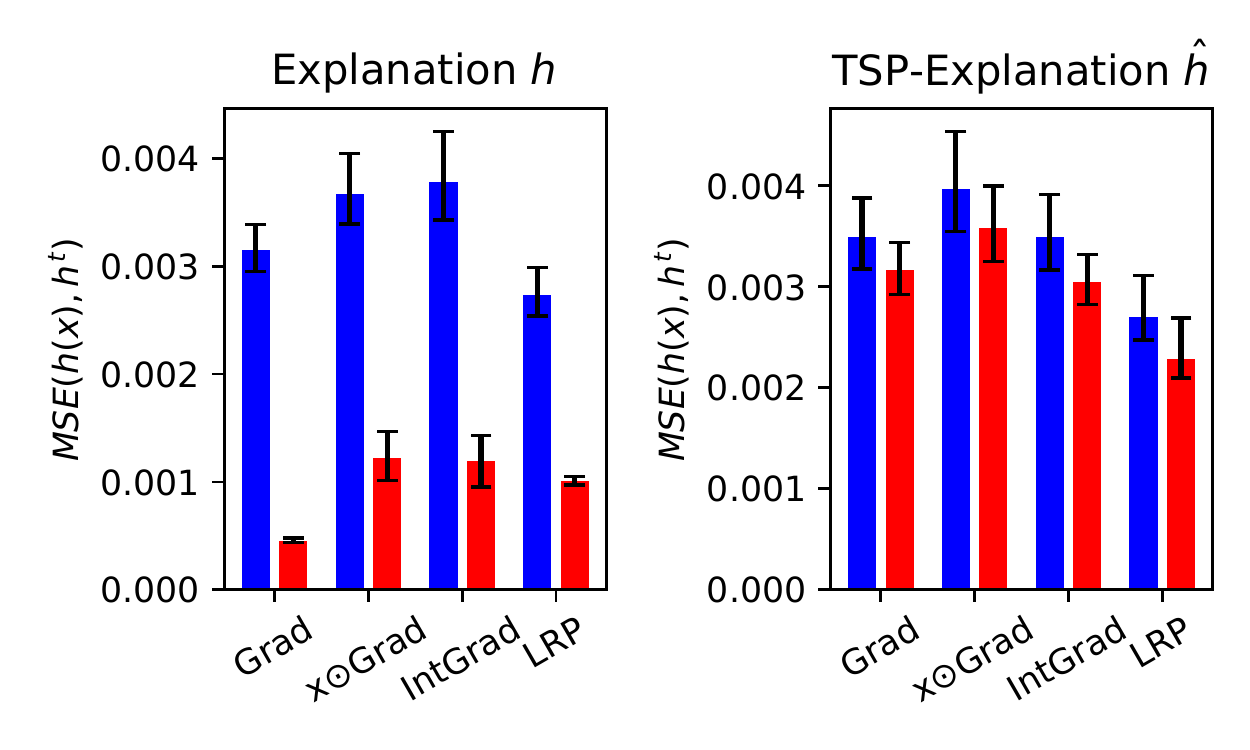}
  \caption{%
      Median of MSE of $\hat{h}_g(x)$ (blue) and $\hat{h}_{\tilde{g}}$ (red) on CIFAR10 where $h(x)$ was manipulated. %
  }
\end{figure}

\begin{figure}[ht!]
  \centering
  \includegraphics[width=1.\linewidth]{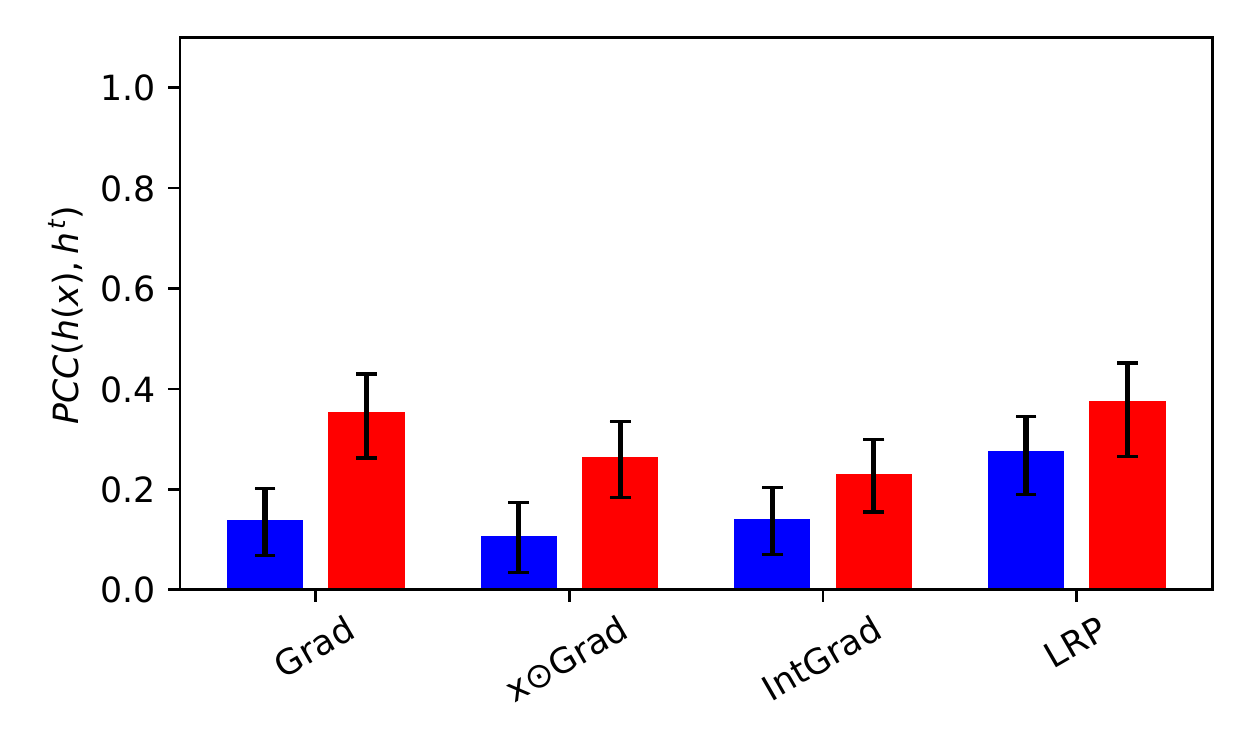}
  \caption{%
      Median of PCC of $\hat{h}_g(x)$ (blue) and $\hat{h}_{\tilde{g}}$ (red) on CIFAR10 where $\hat{h}(x)$ was manipulated. %
  }
\end{figure}

\begin{figure}[ht!]
  \centering
  \includegraphics[width=1.\linewidth]{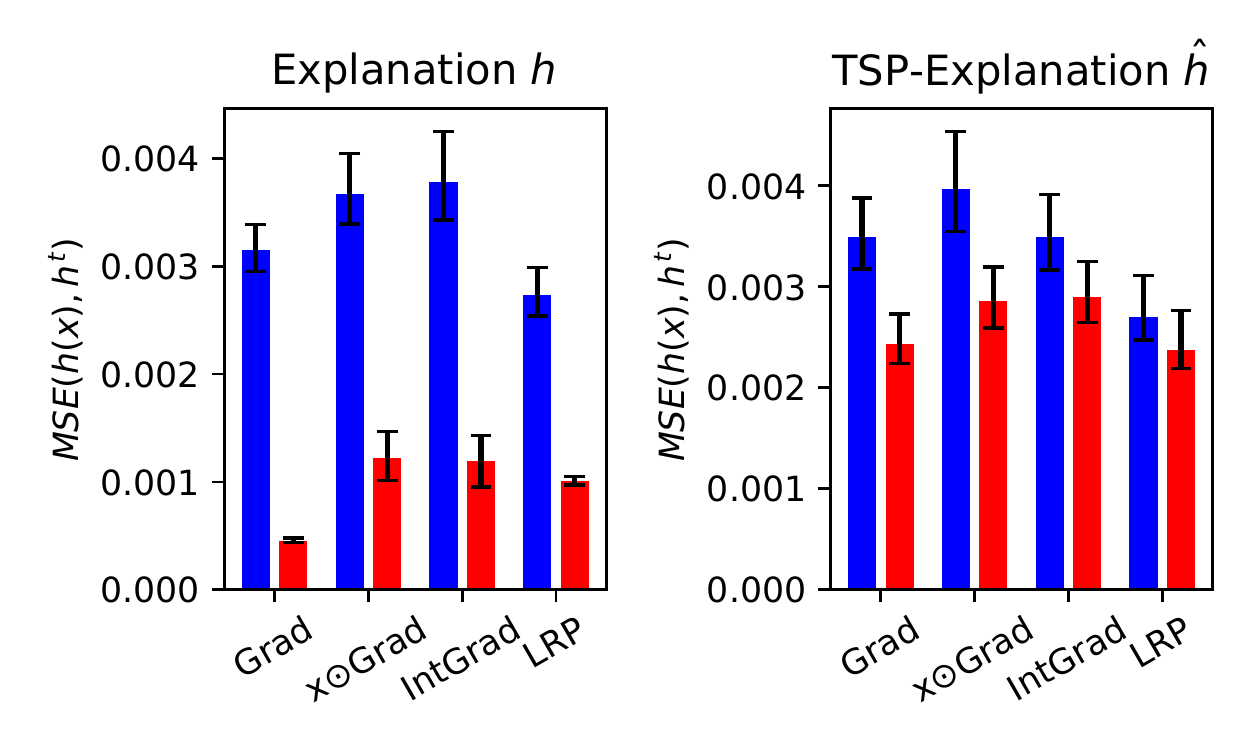}
  \caption{%
      Median of MSE of $\hat{h}_g(x)$ (blue) and $\hat{h}_{\tilde{g}}$ (red) on CIFAR10 where $\hat{h}(x)$ was manipulated. %
  }
\end{figure}

\FloatBarrier
\section{Pixel-flipping}

We compare the original explanations with the respective TSP-explanations using \textit{pixel-flipping}~\citeapp{evaluating}. This metric measures how fast the network confidence $g(x)$ declines when removing features with highest relevance. The pixels are inpainted using the \textit{telea}-method~\citeapp{telea} to alleviate uncontrolled behaviour of the classifier off the manifold. Our result clearly show that tsp-methods perform well on this metric.

\begin{figure}[ht!]
  \centering
  \includegraphics[width=1.\linewidth]{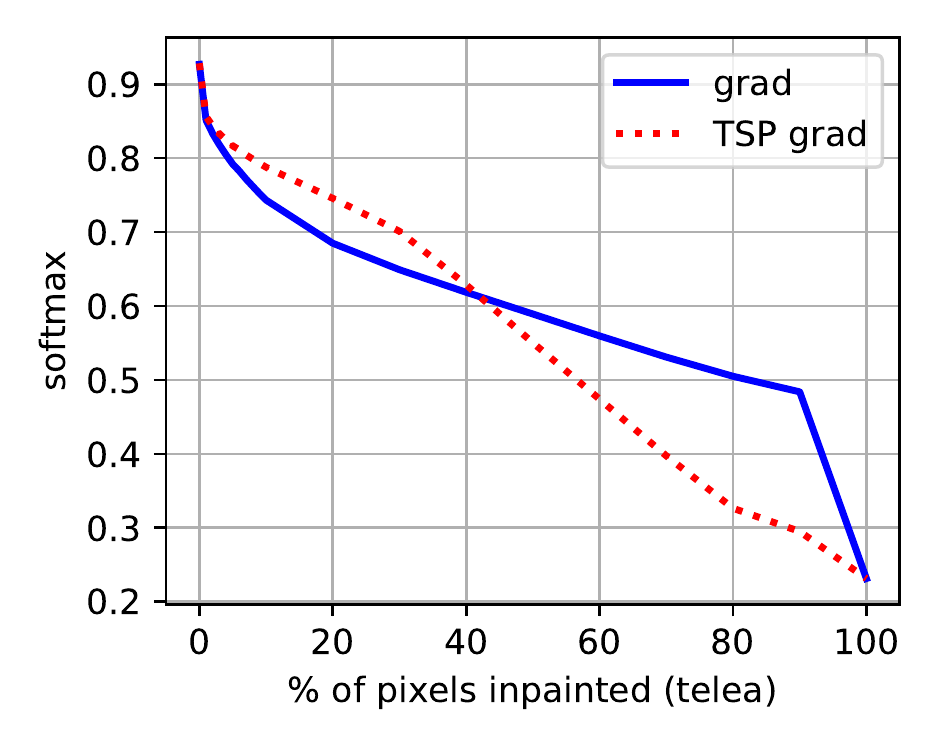}
  \caption{%
      Pixel-flipping performance of Gradient and TSP-Gradient on FashionMNIST %
  }
\end{figure}
\begin{figure}[ht!]
  \centering
  \includegraphics[width=1.\linewidth]{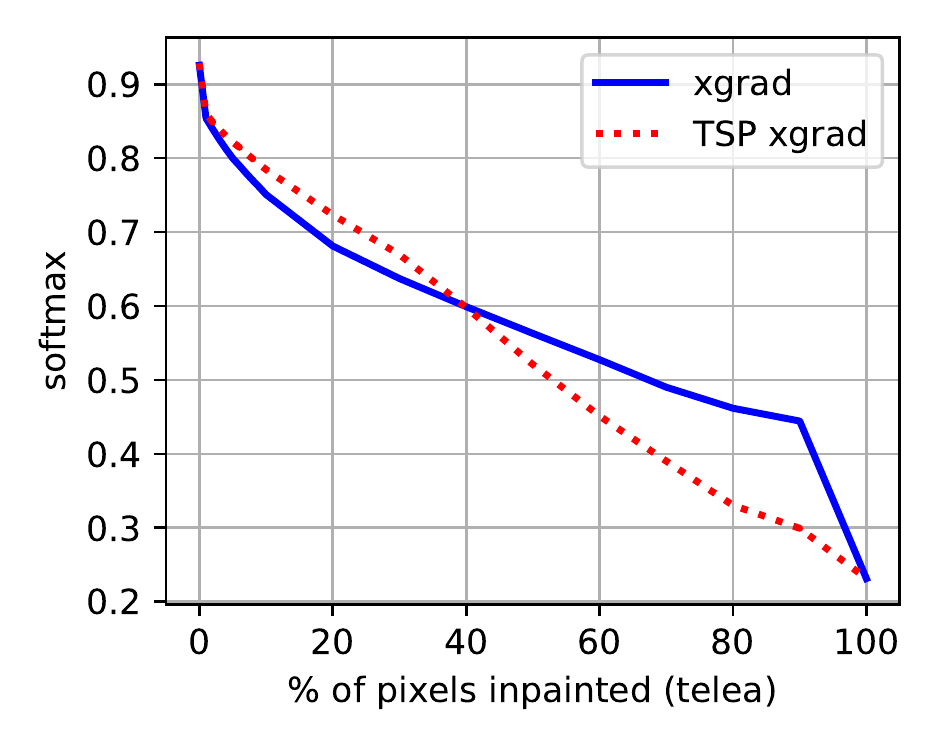}
  \caption{%
      Pixel-flipping performance of x$\odot$Grad and TSP-x$\odot$Grad on FashionMNIST %
  }
\end{figure}
\FloatBarrier

\bibliographystyleapp{icml2020}
\bibliographyapp{refs}

\end{document}